%% file: main.tex
\documentclass{article}

\PassOptionsToPackage{numbers, compress}{natbib}

\usepackage[preprint]{neurips_2023}  %

\usepackage[utf8]{inputenc} 
\usepackage[T1]{fontenc}    
\usepackage{hyperref}       
\usepackage{url}            
\usepackage{booktabs}       
\usepackage{amsfonts}       
\usepackage{nicefrac}       
\usepackage{microtype}      
\usepackage{xcolor}         

\usepackage{array}
\usepackage{algorithm}
\usepackage{algorithmic}
\usepackage{amsthm}
\usepackage{amsmath}
\usepackage{amssymb}
\usepackage{wrapfig}
\usepackage{multirow}
\usepackage{enumitem}
\usepackage{makecell}
\usepackage{graphicx}
\usepackage{appendix}
\usepackage{bm}
\usepackage{soul,color}
\usepackage{adjustbox}

\newtheorem{theorem}{Theorem} 

\newtheorem{prop}{Proposition}

\newtheorem{definition}{Definition}

\title{River of No Return: Graph Percolation Embeddings for Efficient Knowledge Graph Reasoning}

\author{
  Kai Wang, \ \  Siqiang Luo\thanks{Corresponding author}, \ \ Dan Lin\\
  Nanyang Technological University, Singapore\\
  \texttt{{kai\_wang, siqiang.luo, dan.lin}@ntu.edu.sg} \\
}

\begin{document}

\maketitle

\begin{abstract}
  We study Graph Neural Networks (GNNs)-based embedding techniques for knowledge graph (KG) reasoning.
  For the first time, we link the path redundancy issue in the state-of-the-art KG reasoning models based on path encoding and message passing to the transformation error in model training, which brings us new theoretical insights into KG reasoning, as well as high efficacy in practice.
  On the theoretical side, we analyze the entropy of transformation error in KG paths and point out query-specific redundant paths causing entropy increases. These findings guide us to maintain the shortest paths and remove redundant paths for minimized-entropy message passing.
  To achieve this goal, on the practical side, we propose an efficient Graph Percolation Process motivated by the Percolation model in Fluid Mechanics, and design a lightweight GNN-based KG reasoning framework called \textbf{GraPE}. GraPE outperforms previous state-of-the-art methods in both transductive and inductive reasoning tasks, while requiring fewer training parameters and less inference time.
\end{abstract}

\input{sections/introduction.tex}

\input{sections/background.tex}

\input{sections/methodology.tex}

\input{sections/experiments.tex}

\section{Discussion and Conclusions}
\label{sec:6}

\textbf{Limitation and Future Work.}
There are two limitations to GraPE. First, our theoretical analysis is based on the basic path encoding process and the i.i.d. assumption. We will further extend GraPE by designing more refined percolation paths for different subgraphs and GNN architectures.
Second, to process large-scale KGs containing millions or billions of entities, only algorithm improvement of GraPE is not enough, we will further conduct system design and training optimization.

\textbf{Societal Impact.}
Our work drastically reduces the computational time of reasoning models, aiding in the control of carbon emissions. However, as the efficacy of these models improves, their potential for misuse increases, such as exposing sensitive relationships in anonymized data.

\textbf{Conclusions.}
We propose a novel GNN-based KG reasoning framework, Graph Percolation Embeddings (GraPE). 
According to the theoretical analysis of error entropy,
we design a potential-involved path encoding method and extend it to GNN message passing.
GraPE achieves state-of-the-art performance in both transductive and inductive reasoning tasks.
Besides, GraPE has a relatively lower time and space complexity than previous GNN-based methods.

\bibliographystyle{plainnat}
\bibliography{bibtex}

\clearpage
\appendix
\input{sections/appendix.tex}

\end{document}

%% file: sections/introduction.tex
\section{Introduction}

\begin{figure}
\centering
\includegraphics[width=1.0\textwidth]{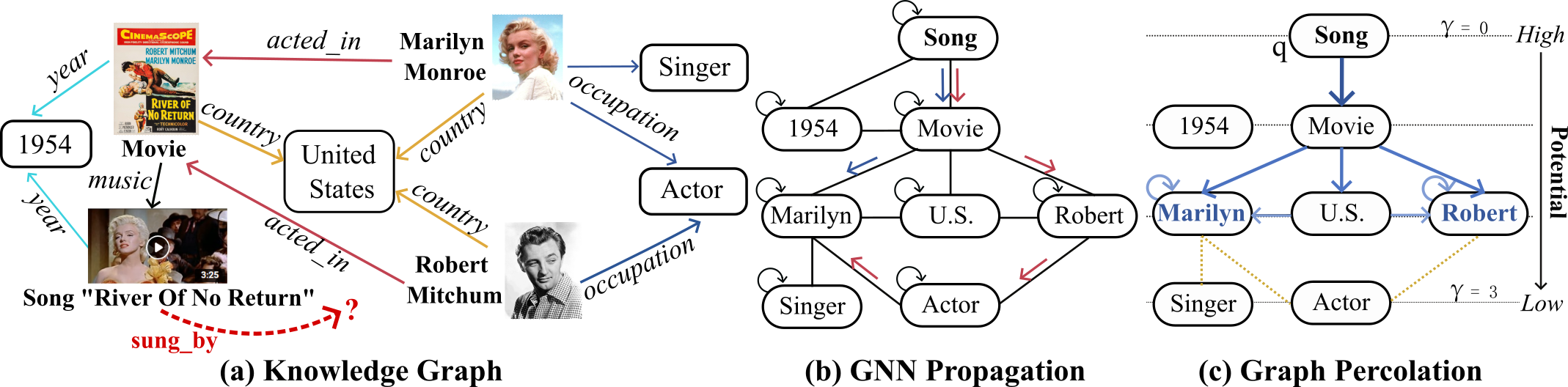}
\vspace{-5mm}
\caption{(a) Example of KG reasoning in a Freebase subgraph. "Marilyn Monroe" is the answer to the query.  (b) A multi-layer GNN propagation with augmented triples traverses all possible paths. (c) The graph percolation only calculates a few triples to encode the entity `Marilyn' and `Robert'. }
\label{fig:1}
\vspace{-5mm}
\end{figure}

The Knowledge Graph (KG) has demonstrated significant potential to organize world knowledge and human commonsense in the form of factual triples (head entity, relation, tail entity) \cite{2020survey,INDIGO-NIPS21,NIPS22-KG1}. 
As KG is not possible to record innumerable world knowledge, there is an increasing research interest in KG reasoning techniques, aiming to deduce new facts from existing KG triples \cite{NIPS21-KG1,OurComp,NIPS21-KG2}. 
The fundamental triple-level KG reasoning task, denoted by the query (head entity, relation, ?), aims to predict the missing tail entity from the entity set of the KG. 
As an example shown in Fig. \ref{fig:1}(a), the answer to the query ("{\em River of No Return}", "{\em sung\_by}") is the entity "{\em Marilyn Monroe}".
One of the mainstream KG reasoning techniques is Knowledge Graph Embedding (KGE) \cite{NIPS22-KG2,NIPS22-KG3,DistMult,OurRotL}. 
Represented by TransE \cite{TransE} and RotatE \cite{RotatE}, KGE models represent entities as $d$-dimensional trainable embedding vectors for further KG reasoning, but suffer from high storage costs and cannot handle unseen entities \cite{OurMulDE,OurHaLE}.

Recent works generate ``relative'' entity embeddings without entity-specific parameters, thereby reducing storage costs and supporting inductive reasoning. 
Path-level methods \cite{PTransE,ComVSM-ACL15,RNNChain-EACL17,PathCon-KDD21} encode entities by aggregating the features along all the paths that reach the candidate entity from the query entity. 
Such path encoding usually processes short paths with up to only three triples, because of high computational costs caused by the exponential growth of paths
To reduce inference complexity, recent subgraph-level methods based on Graph Neural Networks (GNN) further convert complicated path encoding to a subgraph message-passing process in GNNs \cite{GraIL-ICML19,NBF-NIPS21,REDGNN-WWW22}. GraIL \cite{GraIL-ICML19} extracts an enclosing subgraph for each candidate entity. NBFNet \cite{NBF-NIPS21} and RED-GNN \cite{REDGNN-WWW22} propagate the query features layer by layer in the $L$-hop neighborhood subgraph of the query entity. 

However, these state-of-the-art GNN-based methods follow the same GNN message-passing process, i.e. \underline{propagating messages freely through all edges in graphs}, which can bring in redundancy and inefficiency. In particular, the $L$-layer graph propagation in GNNs is equivalent to encoding all possible paths with lengths up to $L$, which is redundant for entities near the query entity to generate relative knowledge embeddings. Meanwhile, the existence of self-loop and reverse edges introduce an exponential number of relational paths, resulting in high computation overhead. For example, in Fig. \ref{fig:1}(b), the dozens of redundant paths like ``Song-Song-1954-Movie-Marilyn'' are computationally intensive in GNNs but contribute negligibly to the query compared with the shortest path ``Song-Movie-Marilyn''.

To overcome the path redundancy issues of traditional graph propagation, we build a novel GNN-based KG reasoning framework, called Graph Percolation Embeddings (\textbf{GraPE}). 
From the novel perspective of the {\em Transformation Error Entropy}, 
we first theoretically analyze the entropy increase caused by path redundancy in the path encoding schema, and derive a clear definition for redundant paths in KG reasoning.
On this basis, we design an efficient {\em Graph Percolation Process} to maintain low-entropy paths and remove redundant paths in the GNN message passing, inspired by the {\em Percolation} process in Fluid Mechanics \cite{pbook1,pbook2}. The illustration of this new paradigm based on entropy-guided percolation is shown in Fig. \ref{fig:1}(c).
After that, a lightweight but effective GNN-based architecture with only two GNN layers is proposed to conduct the multi-layer graph percolation on KG subgraphs. 
To verify the performance of GraPE, we conduct extensive KG reasoning experiments in both transductive and inductive settings on five datasets.
In the transductive reasoning, GraPE outperforms the second-best NBFNet with less than 50\% training parameters and 10\% triple calculations, while in the inductive reasoning, GraPE obtains the best performance using around 30\% parameters and 50\% triple calculations of RED-GNN on average.

The rest of the paper is organized as follows. 
We introduce the background and notations in Section \ref{sec:2}. Section \ref{sec:3} details the GraPE framework and its theoretical analysis. Section \ref{sec:4} reports the experimental studies. Finally, we provide some concluding remarks in Section \ref{sec:6}.

%% file: sections/background.tex
\section{Preliminary}
\label{sec:2}
\subsection{Notations and Definitions}

A Knowledge Graph is in the form of $\mathcal{G} = \{\mathcal{E}, \mathcal{R}, \mathcal{T}\}$, where $\mathcal{T}=\{(e_h,r,e_t)|e_h,e_t \in \mathcal{E},r \in \mathcal{R}\}$ is a set of factual triples, and $\mathcal{E}$, $\mathcal{R}$ are the sets of entities and relations, respectively.
Given a query $(q, r_q)$ containing a query entity $q\in \mathcal{E}$ and a query relation $r_q\in \mathcal{R}$, the knowledge graph reasoning task aims to find the target entity $e_a \in \mathcal{E}$ satisfying that $(q, r_q, e_a)$ or $(e_a, r_q, q)$ belongs to the knowledge graph $\mathcal{G}$.
Following previous work \cite{REDGNN-WWW22, NBF-NIPS21,CompGCN-ICLR20}, we augment the triples in $\mathcal{G}$ with reverse and identity relations. The augmented triple set $\mathcal{T^+}$ is defined as:
$\mathcal{T^+} = \mathcal{T} \cup \{(e_t, r', e_h)|(e_h, r, e_t) \in \mathcal{T}\} \cup \{(e, r_i, e)|e\in  \mathcal{E}\}$,
where relation $r'$ is the reverse relation of a relation $r$, relation $r_i$ refers to the identity relation, and the number of augmented triples is $|\mathcal{T^+}| = 2|\mathcal{T}|+|\mathcal{E}|$.

Furthermore, a relational path from the query entity $q$ to an entity $e_t$ is denoted as $P_{q,e_t}=\{(q, r_1, e_1), (e_1, r_2, e_2), \cdots, (e_{|P|-1}, r_{|P|}, e_t)\}$, which is a set of triples connected head-to-tail sequentially. $|P| \leqslant L$ is the number of triples in the path $P_{e_t}$. 
We define the relative distance $\gamma_{q,e_t}$ as the length of the shortest relational path between $q$ and $e_t$. Unless otherwise specified, we use $P_{e_t}$ and $\gamma_{e_t}$ as the abbreviations of $P_{q,e_t}$ and $\gamma_{q,e_t}$ in this paper.
Then, we donate $\mathcal N_q^\ell$ as the $\ell$-hop neighborhood entities of $q$ in which the relative distance of each entity from $q$ is equal to $\ell$, i.e. $\mathcal N_q^\ell=\{e|\gamma_{e}=\ell,e\in\mathcal{E}\}$.
The $L$-hop neighborhood subgraph $\mathcal{G}_q \subseteq \mathcal{G}$ of the query entity $q$ consists of the triples whose head and tail entities belong to $\mathcal N_q^L$. 
The main notations that will be used in this paper are summarized in Appendix~\ref{app:1}.

\subsection{Related Work: Three Embedding Levels for KG Reasoning}
\label{sec:2.2}

\textbf{Triple-level Absolute Embedding:} 
Traditional entity embedding models, such as TransE \cite{TransE}, DistMult \cite{DistMult}, RotatE \cite{RotatE}, assign an individual, trainable $d$-dimensional vector $\mathbf{e}_i \in \mathbb{R}^d$ for each entity.
The embedding vector of one entity $e_t$ is expected to be close to $\mathbf{e_h} \otimes \mathbf{r}$ for each training triple $ (e_h,r,e_t) \in \mathcal{T^+}$ in the embedding space, such that it can be represented as:
\begin{align}
\mathbf{e}_t = \frac{1}{n}\sum\nolimits_{(e_h,r,e_t) \in \mathcal{T^+}}{(\mathbf{e}_h \otimes \mathbf{r}}), 
\label{eq:kgebasic}
\end{align}
where the transformation operator $\otimes$ transforms the head entity vector $\mathbf{e_h}$ using the relation-specific parameter $\mathbf{r}$. Such absolute embeddings are effective but cannot handle unseen entities after training.

\textbf{Path-level Relative Embedding:} 
Path encoding-based methods \cite{PTransE,ComVSM-ACL15,RNNChain-EACL17,PathCon-KDD21,RNNLogic-ICLR21} aim to capture local entity semantics by encoding relational paths in the KG without entity-specific parameters. 
Given the query entity $q$ as the start node, the basic idea is to represent an entity $e_t$ as the feature aggregation of all relational paths from $q$ to $e_t$ in the $L$-hop neighborhood subgraph $\mathcal{G}_q$, such as:
\begin{align}
\mathbf{e}_{t|q} = \mathcal{F}(\mathcal{P}_{e_t}) = \frac{1}{n}\sum\nolimits_{P \in \mathcal{P}_{e_t}}{(\mathbf{q} \otimes \mathbf{r}_1 \otimes \cdots \otimes \mathbf{r}_{|P|} |_{(e_i,r_i,e'_i) \in P}}),
\label{eq:ikebasic}
\end{align}
where $\mathcal{P}_{e_t}=\{P_{q,e_t}\}$ is the path set and $n$ is the number of relational paths.
However, it is time-consuming since the number of paths grows exponentially w.r.t. path length.

\textbf{Subgraph-level Iterative Embedding:} 
Recent GNN-based methods, such as NBFNet \cite{NBF-NIPS21} and RED-GNN \cite{REDGNN-WWW22}, utilize the iterative process  of graph message-passing to avoid the exponentially-growing triple calculations.
Given $\mathbf{e}_{q|q}^0 = \mathbf{q}$, the formula in the $\ell$-th iteration is as follows:
\begin{align}
\mathbf{e}_{t|q}^{\ell} &= \varphi\Big(\bm{W}^{\ell}\phi\big(\mathbf{e}_{i|q}^{\ell-1} \otimes \mathbf{r}^\ell |_{(e_i, r, e_t)\in\mathcal{G}_{q}}\big), \mathbf{e}_{t|q}^{\ell-1}\Big),
\label{eq:kggnn}
\end{align}
in which the notation $\phi(\cdot)$ refers to the aggregation function that aggregates all triple messages on the $1$-hop neighborhood subgraph, $\bm W^\ell$ is a weighting matrix in the $\ell$-th layer and $\varphi(\cdot)$ is the update function.
Although path encoding in this way can be completed in the polynomial time complexity, this technique still encodes all possible $L$-hop relational paths for each entity, which leads to redundancy.


\subsection{Entropy-Guided Path Redundancy}

In order to reduce redundant paths in the KG reasoning models, we theoretically analyze the path redundancy from the novel view of {\em Transformation Error Entropy}.

\textbf{Transformation Error.} 
Based on the descriptions in Sec. \ref{sec:2.2}, we observe that Equation \eqref{eq:ikebasic} and Equation \eqref{eq:kggnn} can be regarded as a nested form of the triple-level Equation \eqref{eq:kgebasic}, hence path encoding is also based on the same assumption that $\mathbf{e}_t \approx \mathbf{e}_h\otimes\mathbf{r}$. However, when training in the whole KG, trainable parameters would ultimately converge to a sub-optimal solution for each training triple, such that we have $\mathbf{e}_t = \mathbf{e} \otimes \mathbf{r} + \epsilon$. 
We call the error $\epsilon$ between $\mathbf{e}_t$ and $\mathbf{e} \otimes \mathbf{r}$ as the transformation error. Note that, the transformation error objectively exists caused by model training, as long as the model employs a trainable relation vector for one relation $r$ to encode all r-involved triples. 

Considering the transformation error in path encoding, we can measure path redundancy by computing the error entropy. 
A high-entropy path means the message passing through this path entails more uncertainty. For instance, a redundant path ``A-B-A-B-C'' conveys repetitive information and more transformation errors for node C if the path ``A-B-C'' has been calculated. 
Because the transformation error per triple exists but is hard to measure, without loss of generality, we assume that the transformation errors of all triples are independent and identically distributed (i.i.d.). 
Then, our findings about Transformation Error Entropy are formalized in the theorems below: 

\begin{theorem} (Error Propagation)
\label{theo:entropy}
Given the $k$-th triple $(e_{k-1}, r, e_k)$ of a relational path starting from $q$, the transformation error entropy of the relative embedding vector $\mathbf{e}_{k|q}$ is not smaller than that of $\mathbf{e}_{k-1|q}$, i.e., $H(\mathbf{e}_{k|q}) \geqslant H(\mathbf{e}_{k-1|q})$.
\end{theorem}

Theorem \ref{theo:entropy} indicates that transformation errors can propagate through paths and longer relational paths accumulate larger entropy. It motivates us to encode the shortest paths to gather low-entropy entity embeddings.
However, longer paths may contain more semantics and the mean entropy of multiple paths may decrease in Equation \eqref{eq:ikebasic}. 
For instance, path ``A-D-E-C'' has larger entropy than path ``A-B-C'', but the mean entropy of the two paths is lower than each of them. 
Therefore, taking the inevitable entropy increase as the criterion, the definition of redundant paths is clarified as follows:
\begin{definition} (Redundant Path)
\label{def:redpath}
Suppose a path $\hat{P}$ from the query entity $q$ to an entity $e_t$ with its length more than the relative distance $\gamma_{q,t}$, and at least $\gamma_{q,t}$ triples in $\hat{P}$ also exist in the shortest paths from $q$ to $e_t$, then the path $\hat{P}$ is a redundant path for the relative entity embedding $\mathbf{e}_{t|q}$.
\end{definition}
\begin{theorem} (Entropy Increase)
\label{theo:addpath}
Let $\mathcal{P}$ be the set of all shortest paths between the query entity $q$ and a candidate entity $e_t$ in $\mathcal{G}$. If a redundant path $\hat{P}$ is added to the path set $\mathcal{P}$, then the transformation error entropy of the mean-aggregated vector $\mathbf{e}_{t|q}$ would increase, i.e. $H(\mathcal{F}(\mathcal{P} \cup {\hat{P}})) > H(\mathcal{F}(\mathcal{P}))$.
\end{theorem}
Theorem \ref{theo:addpath} indicates that adding a redundant path $\hat{P}$ would inevitably increase the entropy of the mean-aggregated vector of all shortest paths (calculated as Equation \eqref{eq:ikebasic}). 
It is reasonable because the ``effective'' part of $\hat{P}$ is overlapped with existing paths while the longer $\hat{P}$ has larger transformation errors, such as ``A-B-D-B-C'' and ``A-B-A-B-C''.
We prove two theorems in Appendix \ref{app:t1} and \ref{app:t2}. 

In summary, to minimize the transformation error entropy in the path encoding, the above theoretical analysis guides us to solve the path redundancy issue, by maintaining the shortest paths (Theorem \ref{theo:entropy}) and removing redundant paths (Theorem \ref{theo:addpath}).

%% file: sections/methodology.tex
\section{Graph Percolation Embeddings}\label{sec:3}

Integrating all possible paths in path encoding or GNNs requires huge computation costs and accumulates transformation errors.
In order to reduce path amount, previous work \cite{Adaprop,ANet} has attempted to select top K edges in each iteration by random sampling strategies or learnable attention mechanisms, but unavoidably caused information loss resulting from unprocessed triples.
Meanwhile, although Theorem \ref{theo:addpath} points out definite redundant paths, removing them in the GNN message-passing process is still intractable.

\begin{figure*}[!tb]
\centering
\includegraphics[width=0.95\textwidth]{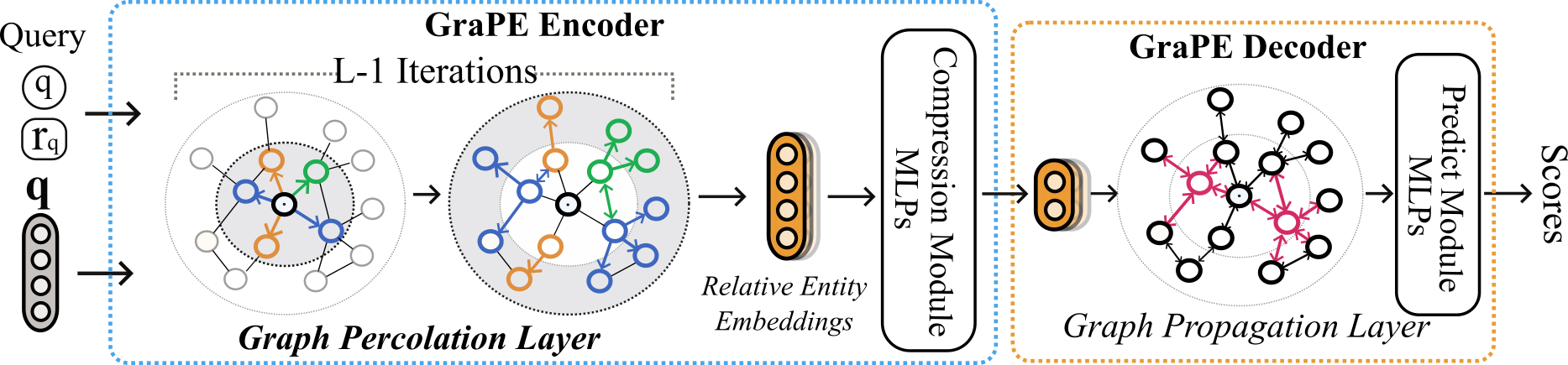}
\vspace{-3mm}
\caption{Graphical illustration of the GraPE architecture.}
\label{fig:3}
\vspace{-5mm}
\end{figure*}

The {\em Percolation} phenomenon in Fluid Mechanics \cite{pbook1,pbook2,pbook3} motivates us to propose a query-specific message-passing process \underline{maintaining shortest paths while avoiding redundant paths without} \underline{complicated calculations and triple loss.} 
Similar to a river flowing downhill through a porous medium under the control of gravity, we model GNN message passing as the graph percolation process by removing ``uphill edges'' to avoid redundant paths, which will be detailed in Sec. \ref{sec:3.1}. On this basis, we propose a novel \textbf{Gra}ph \textbf{P}ercolation \textbf{E}mbeddings (\textbf{GraPE}) framework in Sec. \ref{sec:3.2}. GraPE employs the graph percolation process and achieves efficient Knowledge Graph reasoning in both transductive and inductive settings, whose diagram is shown in Fig. \ref{fig:3}

\subsection{Graph Percolation Process}
\label{sec:3.1}

In Fluid Mechanics, the flow of a fluid through a porous medium is called {\em Percolation}, the percolation model computes the percolation flux following Darcy's law \cite{darcy1856,pbook1}: $\hat{q} = \frac{\kappa \left(p_b-p_a\right)}{\mu h}=\frac{\kappa}{\mu}\Delta h$, where the unit potential difference (or called pressure gradient) is denoted as $\Delta h = (p_b-p_a)/h$. If there is no potential difference over a distance $h$ (i.e. $\Delta h = 0$), no flow occurs. Otherwise, a downhill flow will occur from high potential towards low potential because of gravity or artificial pressure. 

\textbf{Percolation Paths.} Simulating the above gravitational potential in KGs as the potential proportional to the relative distance between each entity and the query entity $q$, it is obvious that a path is the shortest path from $q$ to an entity if and only if each triple $(e_h, r, e_t)$ of the path is ``downhill'' (i.e. $\gamma_{e_h} < \gamma_{e_t}$). Meanwhile, as in Proposition \ref{theo:addedge}, we discover that redundant paths would be introduced by ``uphill'' triples ($\gamma_{e_h} \geqslant \gamma_{e_t}$).

\begin{prop} (See Appendix \ref{app:t3} for proof)
\label{theo:addedge}
Let $\mathcal{\hat{G}}$ be the KG subgraph constructed by the triples of all shortest paths starting from the query entity $q$. Suppose a new triple $t = (e_1, r, e_2)$ whose head and tail entities exist in $\mathcal{\hat{G}}$ and the relative distance $\gamma_{e_1} \geqslant \gamma_{e_2}$. Adding this triple $t$ into $\mathcal{\hat{G}}$ leads to at least one new redundant path passing $t$. 
\end{prop}

To this end, we are motivated to define ``percolation paths'' without uphill triples in the middle. Specifically, given the $\ell$-th triple $(e_1, r, e_2)$ in the path $P (1\leqslant \ell \leqslant |P|)$, the potential difference $\Delta h$ between the two entities are as follows:
\begin{equation}
\Delta h_\ell(e_1,e_2) = \{
\begin{array}{lclcr}
max(\gamma_{e_2} - \gamma_{e_1},0) & & \ell < |P| & &\\
min(\gamma_{e_2} - \gamma_{e_1} + 1, 1) & & \ell = |P| & &
\end{array}
\label{eq:delta}
\end{equation}
\begin{equation}
\mathbf{e}_{t|q} = \frac{1}{n}\sum\nolimits_{P \in \mathcal{P}_{e_t}}{\Big(\Delta h_{|P|}\cdots\Delta h_1(\mathbf{q} \otimes \mathbf{r}_1 \otimes \cdots \otimes \mathbf{r}_{|P|}) |_{(e_i,r_i,e'_i) \in P}}\Big).
\end{equation}
Based on the percolation model, a percolation path is valid only when all the $\Delta h$ values along the path are positive. 
Such that the whole set of percolation paths in a subgraph $\mathcal{G}_q$ satisfies {\em three principles}:
(1) all shortest paths are included;
(2) no redundant path is involved;
(3) all knowledge facts in the subgraph are calculated.
In particular, as defined in Equation \eqref{eq:delta}, the last triple of a percolation path is allowed to be a same-potential triple ($\gamma_{e_2} - \gamma_{e_1} = 0$), because such paths are also not redundant by Definition \ref{def:redpath}. Nevertheless, if we only maintain downhill triples for the shortest paths, the knowledge facts in the same-potential triples would be unavoidably lost.

\textbf{Graph Percolation Process.} We further integrate the percolation paths into the GNN message passing.
Due to the directionality of the percolation paths, the relative embedding vector of an $\ell$-hop entity $e_t \in \mathcal N_{q}^{\ell}$ cannot be influenced by the low-potential entities in the deeper layers. 
Therefore, inspired by the efficiency gained by iterative embeddings in previous work \cite{REDGNN-WWW22}, we combine the percolation paths of all neighborhood entities into $L$ layer-wise subgraphs $\mathcal{\hat{G}}_{q} = \{\{\mathcal{E}_{q}^\ell, \mathcal{R}, \mathcal{T}_{q}'^\ell\} |1 \leqslant \ell \leqslant L\}$ as the following:
\begin{align}
\mathcal{E}_{q}^\ell = \mathcal N_{q}^{\ell-1} \cup \mathcal N_{q}^\ell, \ 
\mathcal{T}_{q}'^\ell = \{(e_1,r,e_2)|e_1\in \mathcal N_{q}^{\ell-1}, e_2\in \mathcal{E}_{q}^\ell\}
\end{align}

\begin{wrapfigure}{R}{0.5\textwidth}
\begin{minipage}{0.5\textwidth}
    \vspace{-2.0em}
    \begin{algorithm}[H]
    	\caption{Graph Percolation Process}
    	\label{alg:gip}
    	\small
    	\begin{algorithmic}[1]
    		\REQUIRE a KG $\mathcal G$, query $(q, r_q)$, and layer number $L$.
    		\STATE initialize $\mathbf{e}_{q|q}^{0}=\mathbf{q}$, $\mathcal{N}_{q}^0 = \{q\}$,
            $\mathbf{e}^0_{t|q}=0$;
    		\FOR{$\ell=1\dots L$}
            \STATE collect the $\ell$-hop entities $\mathcal N_{q}^{\ell}$ from $\mathcal G$;
            \STATE $\mathcal{E}_{q}^\ell = \mathcal N_{q}^{\ell-1} \cup \mathcal N_{q}^\ell$;
            \STATE collect the $\ell$-hop triples $\mathcal{T}_{q}'^\ell$ w.r.t $\mathcal{E}_{q}^\ell$;
            \STATE message passing with Equation \eqref{eq:ginbasic} for $\mathcal{E}_{q}^\ell$;
            \STATE $\mathbf{e}_{t|q}=\mathbf{e}_{t|q}^{\ell}+\mathbf{e}_{t|q}^{\ell-1}$ for entities $e_t\in N_{q}^{\ell-1}$;
    		\ENDFOR
    		\RETURN relative embeddings $\mathbf{e}_{t|q}$ for all $e_t\in\mathcal E$. \label{step:return}
    	\end{algorithmic}
    \end{algorithm}
\end{minipage}
\vspace{-1.0em}
\end{wrapfigure}

Consistent with the percolation paths, the subgraph $\mathcal{\hat{G}}_{q}^\ell$ in the $\ell$-th layer only involve the limited triples from the $(\ell-1)$ hop to the entities in the same hop or the lower $\ell$-th hop.
By integrating the layer-wise subgraphs $\mathcal{\hat{G}}_{q}$ with the basic GNN process in Equation \eqref{eq:kggnn}, the graph percolation process recursively constructs the relative knowledge embeddings of neighborhood entities layer by layer as:
\begin{footnotesize}
\begin{align}
\mathbf{e}_{t|q}^{\ell} &= \delta\Big(\bm{W}^{\ell}\phi\big(\mathbf{e}_{i|q}^{\ell-1} \otimes \mathbf{r}_{i|q}^\ell |_{(e_i, r_i, e_t)\in\mathcal{T}_{q}'^\ell}\big)\Big),
\label{eq:ginbasic}
\end{align}
\end{footnotesize}
where $\mathbf{r}_{i|q}^\ell$ is the relation parameters corresponding to the query relation $r_q$. The entire graph percolation process is shown in Algorithm~\ref{alg:gip}.

It is worth noting that the $L$-layer graph percolation is computationally efficient, because each neighborhood entity only conducts two times of message passing and each triple ($\gamma_{e_h} \leqslant \gamma_{e_t}$) is calculated only once. 
Fig.~\ref{fig:1}(b)(c) vividly illustrate the differences between graph percolation and graph propagation processes. Different from basic graph propagation that considers every triple, graph percolation is like the river of no return, flowing from the source $q$, percolating the entities layer by layer and never going backward.

\subsection{GraPE Architecture}  
\label{sec:3.2}

GNN-based KG reasoning usually requires deep GNNs with at least three layers to cover as many candidate entities as possible. 
As proved in the previous research \cite{GNNDegrad-KDD22}, deep GNNs suffer from over-smoothing and model degradation issues, which would negatively impact the model accuracy. 
Meanwhile, deep GNNs require more trainable parameters, the number of which scales linearly with the number of layers.
In order to improve the efficiency of GNN-based KG reasoning, we design a novel \textbf{GraPE} framework that follows the encoder-decoder architecture, as shown in Fig.~\ref{fig:3}. 

\textbf{Graph Percolation Encoder.} 
Given a query $(q, r_q)$ and the layer-wise subgraphs $\mathcal{\hat{G}}_{q}$, 
the encoder conducts the graph percolation process in Algorithm~\ref{alg:gip} to generate relative knowledge embeddings. 
Different from previous work utilizing multiple GNN layers, the GraPE encoder employs only one GNN layer to process the $L$-layer subgraphs via Equation \eqref{eq:ginbasic}. 
The design space of the relation parameters and other components will be detailed in Appendix \ref{app:md}.

To further reduce complexity, we propose a compression module with two MLP layers to transfer the relative entity embeddings $\mathbf{e}_{t|q}$ from the $d$-dimensional space to a smaller $d_l$-dimensional space as:
$\mathbf{\hat{e}}_{t|q} = \delta\Big(\bm{W_{h2}} \cdot \delta\big(\bm{W_{h1}} \cdot [\mathbf{e}_{t|q}: \mathbf{r}_q]\big)\Big)$,
where $\mathbf{r}_q$ is a trainable relation embedding vector of $r_q$, concatenated with the relative entity vector.
$\bm W_{h1} \in \mathbb{R}^{2d\times d}$ and $\bm W_{h2} \in \mathbb{R}^{d\times d_l}$ are weighting matrices, and $\delta$ is the activation function.

\textbf{Graph Propagation Decoder.}
The GraPE decoder predicts the missing entity using the $d_l$-dimensional relative entity embeddings. 
Here we employ another GNN layer to conduct a basic graph propagation in the whole $L$-hop neighborhood subgraph, which aims to provide the $1$-hop neighbor features to each candidate entity. 
This process is performed once as Equation \eqref{eq:kggnn}, whose time complexity compared with Graph Percolation will be discussed in Sec. \ref{sec:4.3}.
It is necessary because graph percolation ignores the neighbor information on the lower-potential nodes and may fail to distinguish similar candidate entities. For example, in Fig.\ref{fig:3}(c), the entity `Marilyn' and `Robert' would get the same relative embedding vectors if we ignore the specific features of `Singer'.

After that, for each entity $e_t$, GraPE predicts its plausibility score $s_t = \mathcal{H}(\mathbf{\hat{e}}_{t|q}, \mathbf{r}_q)$ via a two-layer MLP module, which is similar to the compression module but compresses embedding vectors from $d_l$ dimensions to one.
Following the previous work \cite{REDGNN-WWW22}, we optimize the GraPE parameters by minimizing the multi-class cross entropy loss with each training triple $(q, r_q, e_a)$.
\begin{align}
\mathcal{L} = \sum\nolimits_{(q, r_q, e_a)\in\mathcal{T'}} \Big( \mathcal{H}(\mathbf{\hat{e}}_{a|q}, \mathbf{r}_q) + \log\big(\!\sum\nolimits_{e_t\in\mathcal E} e^{\mathcal{H}(\mathbf{\hat{e}}_{t|q}, \mathbf{r_q})} \big)
\Big)
\label{eq:loss}
\end{align}
Regardless of the number of layers in the neighborhood subgraph, GraPE only requires two GNN layers for path encoding, i.e. the GNN layer weights for the first $L$$-1$ layers are tied.
Meanwhile, in practice, we conduct only graph percolation in the first $L$$-1$ layers, because the $L$-th layer calculations are overlapped with the decoder propagation.
Furthermore, it is convenient to design different data flows in the two components, better adapting to the local propagation of the shallower layer and the global propagation of the final layer.

%% file: sections/experiments.tex
\section{Experiments}
\label{sec:4}
\subsection{Experimental Setup}

\textbf{Task Settings.} 
To verify the performance of GraPE, we conduct experiments on the KG reasoning tasks.
There are two task settings in current KGh reasoning studies: {\em Transductive Reasoning} and {\em Inductive Reasoning}, which are determined by the scope of the KG facts used to make predictions. 
Specifically, given a knowledge graph $\mathcal{G}_{\text{tra}}=\{\mathcal{E}_{\text{tra}}, \mathcal{R}, \mathcal{T}_{\text{tra}}\}$, 
the transductive KG reasoning task trains and evaluates a model with the same $\mathcal{G}_{\text{tra}}$.
Differently, the inductive KG reasoning task evaluates the trained model on a new knowledge graph $\mathcal{G}_{\text{tst}}=\{\mathcal{E}_{\text{tst}}, \mathcal{R}, \mathcal{T}_{\text{tst}}\}$. $\mathcal{G}_{\text{tra}}$ and $\mathcal{G}_{\text{tst}}$ contain the same set of relations $\mathcal{R}$ but disjoint sets of entities and triples, i.e. 
$\mathcal{E}_{\text{tra}} \cap \mathcal{E}_{\text{tst}}=\emptyset$ and $\mathcal{T}_{\text{tra}} \cap \mathcal{T}_{\text{tst}}=\emptyset$.

\begin{table}[!ht]
	\centering
	\vspace{-5px}
	\caption{Transductive reasoning results on the WN18RR and FB15k237 datasets. 
		The boldface numbers indicate the best performance and the underlined means the second best. }
  \vspace{-5px}
\setlength\tabcolsep{5pt}
\label{tab:tra}
\scriptsize
\begin{tabular}{cl|cccc|cccc}
	\toprule
	\multirow{2}{*}{Type}  &   \multirow{2}{*}{Methods}  &  \multicolumn{4}{c|}{WN18RR}   &  \multicolumn{4}{c}{FB15k237}   \\
	&  &  MRR     & Hits@1  & Hits@3   & Hits@10   &	MRR     & Hits@1  & Hits@3   & Hits@10 \\   \midrule
	\multirow{6}{*}{\makecell[c]{KGE-based}} &  TransE \cite{TransE} & 0.191 & 0.040 & 0.288 & 0.469 & 0.252 & 0.168 & 0.276 & 0.420 \\ 
    &  DistMult \cite{DistMult}     & 0.430 & 0.390 & 0.440 & 0.490	&	0.241 & 0.155 & 0.263 & 0.419 \\ 
    &  ConvE \cite{ConvE}        &  0.430  & 0.440 & 0.440 & 0.520  & 0.325 & 0.237 & 0.356 & 0.501 \\
	&  RotatE  \cite{RotatE}		& 0.477 & 0.428 & 0.492 & 0.571	&	0.337 & 0.241 & 0.375 & 0.533 \\ 
	&  QuatE \cite{QuatE}   	& 0.488 & 0.438 & 0.508 & 0.582 &	0.366 & 0.271 & 0.401 & 0.556	   \\ 
    &  RotH \cite{GoogleAttH}   & 0.496 & 0.449 & 0.514 & 0.586 & 0.344 & 0.246 & 0.380 & 0.535 \\
	\midrule
	\multirow{4}{*}{\makecell[c]{Path-based}} &  MINERVA \cite{MINERVA-NIPS17}  &  0.448 & 0.413 & 0.456 & 0.513	&	0.293 & 0.217 & 0.329 & 0.456  \\ 
    &  NeuralLP \cite{NeuralLP-NIPS17} &0.435& 0.371 & 0.468 & 0.566	&	0.252 & 0.189 & 0.248 & 0.375  \\ 
	&  DRUM \cite{DRUM-NIPS19}  & 0.486 & 0.425 & 0.513 & 0.586	&	0.343 & 0.255 & 0.378 & 0.516	 \\ 
	&  RNNLogic \cite{RNNLogic-ICLR21} & 0.483 & 0.446 & 0.497 & 0.558	&	0.344 & 0.252  & 0.380 & 0.530	 \\ 
	\midrule
	\multirow{6}{*}{\makecell[c]{GNN-based}}  
    &  RGCN \cite{RGCN} & 0.402 & 0.345 & 0.437 & 0.494	&	0.273 & 0.182 & 0.303 & 0.456	 \\ 
	&  CompGCN \cite{CompGCN-ICLR20} &  0.479 & 0.443 & 0.494 & 0.546	&	0.355 & 0.264 & 0.390 & 0.535	 \\ 
	&  RED-GNN  \cite{REDGNN-WWW22} &  0.533 & 0.485 & 0.555 & 0.624& 0.374  & 0.283 & 0.419 & 0.558    \\
	& NBFNet \cite{NBF-NIPS21} & \underline{0.551}   &  \underline{0.497}  &\underline{0.573}&  \underline{0.666}  & \underline{0.415}  &  \underline{0.321}  &\underline{0.454}&  \textbf{0.599}   \\ 
	\cline{2-10}
    & \multirow{2}{*}{\textbf{GraPE}}  &  \textbf{0.570} & 	\textbf{0.517} &	\textbf{0.595} &	\textbf{0.675} &	\textbf{0.425} &	\textbf{0.336} &	\textbf{0.462} &	\textbf{0.599} \\
    & & $\pm$0.001 &	$\pm$0.002 &	$\pm$0.001 &	$\pm$0.003 &	$\pm$0.001 &	$\pm$0.002 &	$\pm$0.002 &	$\pm$0.003 \\
	\bottomrule  
\end{tabular}
\vspace{-5mm}
\end{table}

\textbf{Datasets.} 
Our experimental studies are conducted on five commonly used datasets.
WN18RR \cite{WN18RR} and FB15k237 \cite{FB15k237} are used for the transductive reasoning task. 
The two datasets are extracted from the English lexical database WordNet and the knowledge base Freebase, respectively. 
For the inductive reasoning task, we use the three series of benchmark datasets \cite{GraIL-ICML19} created on WN18RR \cite{WN18RR}, FB15k237 \cite{FB15k237} and NELL-995 \cite{NELL995}. 
The statistics of the datasets are given in Table \ref{table:ds-tra} and Table \ref{table:ds-ind} in Appendix.
We use two evaluation metrics for both task settings following the previous work \cite{TransE,GraIL-ICML19}. MRR (Mean Reciprocal Rank) is the average inverse rank of test triples and
Hits@N is the proportion of correct entities ranked in the top N.
A higher MRR and Hits@N indicate improved performance.

\textbf{Implementation Details.} 
We select the hyperparameters of our model via grid search according to the metrics on the validation set. For the model architecture, we set the default transform operator $\otimes$ as the Hadamard product in DistMult \cite{DistMult} and the aggregation function $\phi(\cdot)$ as the PNA aggregator \cite{PNA-NIPS20}. Following the data preprocessing of NBFNet \cite{NBF-NIPS21}, we drop out triples that directly connect query entities during training on FB15k237.
More hyperparameter configurations on different datasets are shown in Appendix \ref{app:md}.
All experiments are performed on Intel Xeon Gold 6238R CPU @ 2.20GHz and NVIDIA RTX A5000 GPU, and are implemented in Python using the PyTorch framework.

\subsection{Main Experiments}


\textbf{Transductive KG Reasoning.}
We compare GraPE with 14 baselines, including six KGE-based, four path-based, and four GNN-based models. The experimental results on WN18RR and FB15k237 are shown in Table \ref{tab:tra}.
We observe that GraPE outperforms existing methods on all metrics of the two datasets.
Compared with traditional KGE-based and path-based baselines, GraPE achieves significant performance gains. Especially, the MRR of GraPE has a more than 10\% increase on both datasets which improves from 0.496 to 0.568 on WN18RR and from 0.366 to 0.423 on FB15k237.

In the five GNN-based models, we find that two models based on the absolute knowledge embeddings, RGCN and CompGCN, are significantly weaker than the latter three. It proves the effectiveness of the relative knowledge embeddings on KG reasoning.
Benefiting from the filtered relation paths and lower transformation errors,
GraPE outperforms the state-of-the-art NBFNet, especially on MRR and Hits@1. It indicates that the relative embeddings of GraPE can better distinguish similar entities and achieve accurate predictions. 
Although the performance gains in Hit@3 and Hit@10 are smaller than 1\%, the computational complexity and model cost of GraPE are much lower than NBFNet, which will be discussed in Sec. \ref{sec:4.3}.

\begin{table}[!ht]
	\centering
    \vspace{-3mm}
	\caption{Inductive reasoning results on three series of datasets (evaluated with MRR).}
	\label{tab:ind}
	\scriptsize
	\setlength\tabcolsep{5pt}
	\begin{tabular}{l|cccc|cccc|cccc}
		\toprule
		    \multirow{2}{*}{Methods}   & \multicolumn{4}{c|}{WN18RR} & \multicolumn{4}{c|}{FB15k237}  & \multicolumn{4}{c}{NELL-995} \\
	   & V1    & V2   & V3   & V4   & V1    & V2    & V3    & V4    & V1  & V2  & V3  & V4  \\   
		\midrule
	  RuleN \cite{RuleN-2018}  &  .668   & .645  & .368  & .624  &  .363  & .433  &  .439  & .429  & .615 &  .385& .381 &  .333\\
		NeuralLP \cite{NeuralLP-NIPS17} &	.649	&	.635	&	.361	&	.628	&	.325	&	.389	&	.400	&	.396	&	.610	&   .361 & .367  &   .261	\\
		DRUM  \cite{DRUM-NIPS19}    &  .666  &  .646    &   .380   &   .627   &   .333    &  .395     &   .402    &   .410    &  .628  &  .365  & .375  &  .273  \\
		\cline{1-13}
		GraIL \cite{GraIL-ICML19}    &    .627    &   .625   &   .323   &  .553    &  .279     &   .276    &    .251   &    .227  &  .481 & .297 &  .322  & .262  \\
  	NBFNet \cite{NBF-NIPS21}     &    .685 & 	.659 & 	.417 & 	.610 & 	.306 & 	.344 & 	.328 & 	.312 & 	.481 & 	.379 & 	.385 & 	.203 \\ 
		{RED-GNN \cite{REDGNN-WWW22}}      &   \underline{.701}  &  \underline{.690}   &  \underline{.427}    &   \underline{.651}   &     \underline{.369}     &  \underline{.469}     &  \underline{.445}     &  \underline{.442}  &  \underline{.637}   &  \underline{.419} &   \underline{.436}  &  \underline{.363}   \\    
		\cline{1-13}
        \textbf{GraPE}  &  \textbf{.742}	&  \textbf{.707}	&  \textbf{.472}	&  \textbf{.653}	&  \textbf{.415}	&  \textbf{.488}	&  \textbf{.481}	&  \textbf{.470}	&  \textbf{.777}	&  \textbf{.494}	&  \textbf{.450}	&  \textbf{.383} \\
        $(\pm)$ &	.007	&	.003	&	.006	&	.003	&	.006	&	.007	&	.009	&	.006	&	.012	&	.004	&	.011	&	.011 \\
		\bottomrule
	\end{tabular}
 \vspace{-3mm}
\end{table}

\textbf{Inductive KG Reasoning.}
Considering most previous models cannot handle the inductive settings, we compare GraPE against three path-based and three GNN-based baselines and summarize the MRR results on three series of inductive datasets in Table \ref{tab:ind}. 

From Table \ref{tab:ind}, we have the following observations. 
On all inductive subsets of three datasets, GraPE achieves state-of-the-art performance.
Compared with the previous best method RED-GNN, GraPE obtains an average 8\% relative performance gain in MRR. Especially, the MRR of GraPE improves from 0.369 to 0.422 on the FB15k237-v1 dataset, and from 0.419 to 0.517 on NELL-995-v2. In the inductive task, the generalization ability of relational features learned from paths is much more important than that in the transductive one. Instead of encoding all possible relational paths in RED-GNN, GraPE only extracts features from the limited low-entropy paths, thus improving precision and efficiency simultaneously.

\subsection{Ablation Studies}

We further verify the GraPE performance with different numbers of layers and dimensions and different GNN functions. The experimental results evaluated with MRR are shown in Table \ref{tab:abl}. 

\begin{wraptable}{r}{0.5\textwidth}
    \vspace{-5mm}
    \caption{Ablation studies of GraPE on four datasets (evaluated with MRR).}
    \label{tab:abl}
    \scriptsize
    \resizebox{0.5\textwidth}{!}{
    \begin{tabular}{p{20pt}p{20pt}cp{18pt}p{18pt}p{18pt}}
        \toprule
        \multicolumn{2}{c}{\textbf{Model}}  & \textbf{WN18RR} & \textbf{WNv1} & \textbf{FBv1} & \textbf{NEv1} \\
        \midrule
        \multicolumn{2}{l}{origin}	& \textbf{0.570}	& \textbf{0.742}	& \textbf{0.415} & \textbf{0.777}  \\
        \multicolumn{2}{l}{(a) w all paths}	& 0.570	& 0.735	& 0.396	& 0.769  \\
        \multicolumn{2}{l}{(b) w shortest paths} &	0.565	& 0.741	& 0.401	& 0.634  \\
        \multicolumn{2}{l}{(c) w long paths} &	0.534	& 0.516	& 0.290	& 0.657  \\
        \multicolumn{2}{l}{(d) w/o decoder} &	0.548	& 0.725	& 0.385	& 0.726  \\
        \multicolumn{2}{l}{(e) w/o rel feature} &	0.566	& 0.740	& 0.398	& 0.741  \\
        \midrule
        \multicolumn{6}{c}{\textbf{(a) Different GraPE variants}}   \\
        \midrule
        \textbf{Dim($d$)} & \textbf{Layer($L$)} & \textbf{WN18RR} & \textbf{WNv1} & \textbf{FBv1} & \textbf{NEv1} \\
        \midrule
        32           & 4  & 0.548       & 0.739       & 0.415       & 0.696       \\
        32           & 5  & 0.561       & 0.742       & 0.412       & 0.777       \\
        32           & 6  & 0.562       & 0.741       & 0.412       & 0.759       \\
        64           & 4  & 0.550       & \textbf{0.753}          & \textbf{0.416}          & 0.768       \\
        64           & 5  & \textbf{0.570}          & 0.750       & 0.415       & 0.683       \\
        64           & 6  & 0.567       & 0.741       & 0.404       & \textbf{0.778}         \\
        \midrule
        \multicolumn{6}{c}{\textbf{(b) Different layers and dimensions}}            \\
        \midrule
        \textbf{AGG}$\phi$ & \textbf{TRA}$\otimes$ & \textbf{WN18RR} & \textbf{WNv1} & \textbf{FBv1} & \textbf{NEv1} \\
        \midrule
        PNA          & DistMult     & \textbf{0.570}          & \textbf{0.742}          & \textbf{0.415}          & \textbf{0.777}          \\
        MEAN         & DistMult     & 0.545       & 0.731       & 0.403       & 0.702       \\
        SUM          & DistMult     & 0.542       & 0.718       & 0.376       & 0.743       \\
        PNA          & TransE       & 0.559       & 0.740       & 0.395       & 0.736       \\
        PNA          & RotatE       & 0.561          & 0.739       & 0.400       & 0.726      \\
        \midrule
        \multicolumn{6}{c}{\textbf{(c) Different GNN functions}}   \\
        
    \end{tabular}
    }
    \vspace{-5mm}
\end{wraptable}

\textbf{Comparison of GraPE Variants.} Table \ref{tab:abl}(a) compares different variants of GraPE. Variant (a) utilizes all paths without the percolation function. Variants (b)(c) utilize different parts of paths in the percolation encoder. Variant (d) has an additional percolation layer to replace the decoder GNN layer, while variant (f) does not input relation vectors into MLP layers.
Compared to variant (a) using all paths, GraPE achieves higher MRR metrics on three of four datasets. It proves the effectiveness of the graph percolation process. Variants (b)(c) show that encoding the shortest paths obviously outperforms using long paths. The performance declines of the two variants (d)(f) also indicate the effectiveness of the corresponding components.

\textbf{Subgraph Layers \& Embedding Dimensions.} Table \ref{tab:abl}(b) compares the results of GraPE with different numbers of layers and embedding dimensions. 
The results with a four-layer subgraph on WN18RR are significantly lower than those with more layers, because more subgraph layers can cover more candidate entities in the neighborhood. 
In contrast, the KGs on the inductive datasets usually have fewer entities, and sometimes a four-layer propagation already covers the whole entity set. 
Although the 64-dimensional GraPE outperforms the lightweight one in several metrics, the performance gains are not significant. With the same subgraph layers, the 64-dimensional GraPE only obtains an average 1\% relative performance gain in MRR. Considering the model inference time and GPU memory costs, we recommend using the 32-dimensional setting in inductive reasoning to balance both accuracy and efficiency.

\textbf{Core Functions in GNN Layers.} Table \ref{tab:abl}(c) compares the results of GraPE with different transform operators and aggregation functions. Compared with the PNA aggregation containing multiple aggregators and scalars, the pure SUM and Mean aggregations perform weaker on the four datasets, especially on WN18RR. Meanwhile, using vector addition in TransE or vector rotation in RotatE as the transform operator has relatively fewer performance changes. Overall, the group of the PNA aggregation and Hadamard product in DistMult achieves better and more robust performance on all datasets.

\subsection{Complexity Analysis} 
\label{sec:4.3}

Benefiting from the graph percolation and the lightweight architecture, GraPE achieves a lower time and space complexity than previous GNN-based methods and requires less inference time than the previous two methods. 
In Appendix \ref{app:cc}, we compare the computational complexity of GraPE and previous methods. 

\begin{wraptable}{r}{0.5\textwidth}
    \vspace{-7mm}
    \caption{Complexity comparison on training parameters, involved triples, and inference time. \textbf{$\times R$} denotes the ratio between the right method and GraPE. Bigger \textbf{$R$} means higher complexity.}
    \label{tab:eff}
    \scriptsize
    \resizebox{0.5\textwidth}{!}{
    \begin{tabular}{p{38pt}p{18pt}p{42pt}p{10pt}p{28pt}p{10pt}} 
    \toprule
    \textbf{Datasets}     & \textbf{GraPE} & \textbf{RED-GNN} & \textbf{$\times R$} & \textbf{NBFNet} & \textbf{$\times R$} \\
    \midrule
    \textbf{WN18RR}       & 45,657         & 57,719          & 1.3            & 89,601          & 2.0            \\
    \textbf{FB15k237}     & 110,745        & 117,500         & 1.1            & 3,103,105       & 28.0           \\
    \textbf{WN-v1}   & 12,793         & 56,439          & 4.4            & 88,705          & 6.9            \\
    \textbf{FB-v1} & 40,153         & 58,636          & 1.5            & 2,377,153       & 59.2           \\
    \textbf{NE-v1}  & 13,593         & 59,639          & 4.4            & 90,945          & 6.7            \\
    \midrule
    \multicolumn{6}{c}{\textbf{(a) Training Parameters}}            \\
    \midrule
    \textbf{Datasets}     & \textbf{GraPE} & \textbf{RED-GNN} & \textbf{$\times R$} & \textbf{NBFNet} & \textbf{$\times R$} \\
    \midrule
    \textbf{WN18RR}       & 10,358         & 11,764          & 1.1            & 162,223         & 15.7           \\
    \textbf{FB15k237}     & 300,002        & 823,304         & 2.7            & 3,351,898       & 11.2           \\
    \textbf{WN-v1}   & 239            & 388 & 1.6            & 1,742           & 7.3            \\
    \textbf{FB-v1} & 1,273          & 1,481           & 1.2            & 19,433          & 15.3           \\
    \textbf{NE-v1}  & 2,720          & 6,414           & 2.4            & 11,346          & 4.2            \\
    \midrule
    \multicolumn{6}{c}{\textbf{(b) Involved Triples}}       \\
    \midrule
    \textbf{Datasets}     & \textbf{GraPE} & \textbf{RED-GNN} & \textbf{$\times R$} & \textbf{NBFNet} & \textbf{$\times R$} \\
    \midrule
    \textbf{WN18RR}       & 24s         & 33s          & 1.4            & 120s         & 5           \\
    \textbf{FB15k237}     & 120s        & 1,920s         & 16            & 1,320s       & 11           \\
    \textbf{WN-v1}   & <1s            & 1s & 1            & 2s           & 2            \\
    \textbf{FB-v1} & <1s          & 2s           & 2            & 1s          & 1           \\
    \textbf{NE-v1}  & <1s          & 3s           & 3            & 1s          & 1            \\
    \bottomrule
    \multicolumn{6}{c}{\textbf{(c) Inference Time}}  
    \end{tabular}}
\vspace{-7mm}
\end{wraptable}

We further compare the actual parameter amounts of three GNN-based methods when they achieve state-of-the-art performance.
Table \ref{tab:eff}(a) shows the parameter amount on five datasets. All data is calculated based on the best hyperparameter settings reported by these methods.
Because of the lightweight architecture, GraPE requires the fewest parameters to get the best performance on all datasets.
Comparing GraPE and RED-GNN, we find that 
RED-GNN requires much more parameters on the WN-v1 and NE-v1 datasets. It is because the number of vector dimensions $d$ required by RED-GNN on the two datasets is two times larger than that of GraPE. While GraPE still gets better performance with fewer parameters. 
Meanwhile, NBFNet requires much more parameters than the other two methods, because it utilizes a linear layer for relation embeddings with $O\big(L|\mathcal{R}|d^2\big)$ costs. 
Therefore, adopting NBFNet on complicated KGs is very resource-intensive.

We also compare the total computations of triples involved by the three state-of-the-art methods in the inference period. The average number of triples that a model calculates for one query is shown in Table \ref{tab:eff}(b). 
We observe that GraPE needs the fewest triple calculations benefiting from the graph percolation.
NBFNet encodes all triples in the subgraph with $L$ times, hence it costs the highest calculations. In the two datasets of transductive tasks, GraPE only calculates ten percent of triples and obtains better performance than NBFNet.
Meanwhile, RED-GNN encodes triples in the $\ell$-hop with $(L-\ell+1)$ times in order to encode all relational paths. The redundant paths negatively influence the embedding effect and force RED-GNN to do reasoning with more layers.   

We further compare the inference time of three GNN-based methods. 
The running time per inference epoch on the validation dataset is shown in Table \ref{tab:eff}(c). 
We observe that GraPE achieves much less inference time than RED-GNN and NBFNet on both transductive datasets. Especially on FB15k-237, NBFNet gets the second-best performance requiring a ten times longer inference phase.
The efficiency differences are relatively small on the three inductive datasets, because of the small graph scale.

In summary, GraPE outperforms the second-best NBFNet with less than 50\% parameters and 10\% triple calculations in the transductive task, while in the inductive task, GraPE gets the best performance using around 30\% parameters and 50\% triple calculations of RED-GNN on average.

%% file: sections/appendix.tex
\section{Summary of Notations} 
\label{app:1}
The main notations used in this paper and their descriptions are summarized in Table~\ref{tab:app1}.

\begin{table}[h]
\caption{Summary of the major notations in this paper.}
\centering
\label{tab:app1}
\begin{tabular}{l|l}
\hline
\textbf{Symbol} & \textbf{Description} \\[3pt]
\hline
$\mathcal{G}$ & A knowledge graph (KG) \\[3pt]
$\mathcal T$ & The set of existing triples in a KG\\[3pt]
$\mathcal T'$ & The set of augmented triples\\[3pt]
$\mathcal E,\mathcal R$ & The entity set and relation set in a KG\\[3pt]
$|\mathcal T|, |\mathcal E|$ & The item number in a specific set\\ [3pt]
$e,r$ & An entity (e) or a relation (r) in a KG\\[3pt]
$(q,r_q)$ & A query with an entity $q$ and a relation $r_q$ \\[3pt]
$\mathcal{G}_{q}\subseteq\mathcal{G}$ & The neighborhood subgraph of the entity $q$ \\[3pt]
$\mathbf{e}_i$ & The absolute embedding vector of the entity $e_i$\\[3pt]
$\mathbf{e}_{i|q}$ & The relative embedding vector of $e_i$ relative to $q$\\[3pt]
$\mathbf{r}$ & The parameters corresponding to the relation $r$\\[3pt]
$d, d_l$ & Dimension of embedding vectors or features\\[3pt]
$L$ & Number of GNN layers or neighborhood hops\\[3pt]
$\mathcal{P}_{q,e_t},P_{q,e_t}$ & The path set and a relational path from $q$ to $e_t$\\[3pt]
$\mathcal{F}(\mathcal{P}_{e_t})$ & The path encoding process with the path set $\mathcal{P}_{e_t}$\\[3pt]
$\gamma_{q,e_t},\gamma_{e_t}$ & The relative distance from $q$ to $e_t$\\[3pt]
$\mathcal N_{q}^\ell$ & The $\ell$-th hop neighbors of $q$\\[3pt]
$H(\epsilon)$ & The information entropy of the error $\epsilon$ \\[3pt]
$\Delta h (e_1, e_2)$ & The potential difference between two entities \\[3pt]
$\mathbf{e_h} \otimes \mathbf{r}$ & The transform operator of a KGE model\\[3pt]
$\phi(\cdot)$ & The aggregation function of a GNN model \\[3pt]
\hline
\end{tabular}
\end{table}

\section{Proof} 
\subsection{Proof for Theorem \ref{theo:entropy}} 
\label{app:t1}

{\em 
Given the $k$-th triple $(e_{k-1}, r, e_k)$ of a relational path starting from $q$, the transformation error entropy of the relative embedding vector $\mathbf{e}_{k|q}$ is not smaller than that of $\mathbf{e}_{k-1|q}$, i.e., $H(\mathbf{e}_{k|q}) \geqslant H(\mathbf{e}_{k-1|q})$.
}

\begin{proof}   
In most embedding-based models, affine transformations are employed to model the interactions between the relative embedding vector of the head entity and the relation-specific parameter within a triple. 
Given the triple $(e_{k-1}, r, e_k)$ in the path, we denote the general form of a relational affine transformation as follows:
\begin{align}
\mathbf{e}_{k|q} = \mathbf{e}_{k-1|q} \otimes \mathbf{r}_k + \mathbf{\epsilon}  = \mathbf{A}_k \mathbf{e}_{k-1|q} + \mathbf{b}_k+ \mathbf{\epsilon},
\end{align}
where $\mathbf{A}_k \in \mathcal{R}^{d\times d}$, $\mathbf{b}_k \in \mathcal{R}^{d}$ are the linear part (matrix) and the translation part (vector) of the affine transformation. Without loss of generality, we assume that the transformation error $\mathbf{\epsilon}_k$ in each triple is independent and identically distributed Gaussian noise, i.e.: $\epsilon \sim \mathcal N (0, \hat{\sigma}^2)$. Then, we can represent the variance of the entity vector $\mathbf{e}_{k|q}$ as:
\begin{align}
\text{Var}(\mathbf{e}_{k|q}) = \text{Var}(\mathbf{A}_k \mathbf{e}_{k-1|q} + \mathbf{b} + \mathbf{\epsilon}) = \mathbf{A}_k \text{Var}(\mathbf{e}_{k-1|q}) \mathbf{A}_k^T + \hat{\sigma}^2.
\label{euqa:var0}
\end{align}
Meanwhile, we already know the information entropy of Gaussian distribution $H(\mathcal N (\mu, \sigma^2)) = \frac{1}{2}\ln \left(2 \pi e \sigma^2\right)$. In this formula, when the variance $\sigma^2$ increases, the information entropy will also increase accordingly. Therefore, we have:
\begin{align}
H(\mathbf{e}_{k|q}) \geqslant H(\mathbf{e}_{k-1|q}) &\Leftrightarrow \text{Var}(\mathbf{e}_{k|q}) \geqslant \text{Var}(\mathbf{e}_{k-1|q}) \\
&\Leftrightarrow \mathbf{A}_k \text{Var}(\mathbf{e}_{k-1|q}) \mathbf{A}_k^T + \hat{\sigma}^2 \geqslant \text{Var}(\mathbf{e}_{k-1|q}).
\label{euqa:var}
\end{align}

To prove Equation \eqref{euqa:var}, we discuss three major types of affine transformations in KGE models:
\begin{itemize}
    \item [\normalsize{\textcircled{\scriptsize{1}}}] Translation Transformation (used in TransE \cite{TransE}):  $\mathbf{A}_k$ is the identity matrix ($\mathbf{I}$).
    \item [\normalsize{\textcircled{\scriptsize{2}}}] Scaling Transformation (used in DistMult \cite{DistMult}): $\mathbf{A}_k$ is a diagonal matrix.
    \item [\normalsize{\textcircled{\scriptsize{3}}}] Rotation Transformation (used in RotatE \cite{RotatE}): $\mathbf{A}_k$ is an orthogonal matrix.
\end{itemize}

For the types \normalsize{\textcircled{\scriptsize{1}}} and \normalsize{\textcircled{\scriptsize{3}}}, the $\mathbf{A}_k$ is orthogonal ($\mathbf{A}_k \mathbf{A}_k^T=\mathbf{I}$). Because the orthogonal matrix $\mathbf{A}_k$ preserves the variances in all directions when transforming $\text{Var}(\mathbf{e}_{k-1|q})$, we have $\mathbf{A}_k \text{Var}(\mathbf{e}_{k-1|q}) \mathbf{A}_k^T \geqslant \text{Var}(\mathbf{e}_{k-1|q})$. Therefore, Equation \eqref{euqa:var} holds.

For the type \normalsize{\textcircled{\scriptsize{2}}}, it is complicated to analyze each relation-specific diagonal matrix, so we assume one general diagonal matrix $\mathbf{A} = \text{diag}(a_1,a_2,\cdots, a_d)$ for k steps. 
And the single item of embedding variance $(\mathbf{A} \text{Var}(\mathbf{e}_{k-1|q}) \mathbf{A}^T)_{i} = a_{i}^2 \sigma_{(k-1,i)}^2$. As the embedding error is Gaussian distributed, Equation \eqref{euqa:var} can be represented as:
\begin{align}
\sum_{i=0}^d \sigma_{(k,i)}^2 = \sum_{i=0}^d a_{i}^2 \sigma_{(k-1,i)}^2 + \sum_{i=0}^d\hat{\sigma}_{i}^2 \geqslant \sum_{i=0}^d \sigma_{(k-1,i)}^2 \Leftarrow a_{i}^2 \sigma_{(k-1,i)}^2 + \hat{\sigma}_{i}^2 \geqslant \sigma_{(k-1,i)}^2,
\label{euqa:var2}
\end{align}
where $\sigma_{(k-1,i)}^2$ and $\hat{\sigma}_{i}^2$ are the $i$-th dimentional variance of $\text{Var}(\mathbf{e}_{k-1|q})$ and $\hat{\sigma}^2$.

Therefore, we further prove Equation \eqref{euqa:var2} for the type \normalsize{\textcircled{\scriptsize{2}}} using mathematical induction:

Base case (k=1): The variance of the query vector $\text{Var}(\mathbf{q})$ is zero, i.e. $\sigma_{(0,i)}^2=0$, and $\hat{\sigma}_{i}^2 \geqslant 0$. Thus, Equation \eqref{euqa:var2} holds for $k = 1$.

Inductive step: Assume that Equation \eqref{euqa:var2} is true for $k = n$, where $n$ is an arbitrary natural number. That is, we assume that:
\begin{align}
a_{i}^2 \sigma_{(n-1,i)}^2 + \hat{\sigma}_{i}^2 \geqslant \sigma_{(n-1,i)}^2 \Leftrightarrow (a_{i}^2 - 1) \sigma_{(n-1,i)}^2 + \hat{\sigma}_{i}^2 \geqslant 0.
\label{euqa:var3}
\end{align}    

Now, we aim to show that Equation \eqref{euqa:var2} is also true for $k = n+1$:
\begin{align}
&\text{Var}(\mathbf{e}_{n+1|q})-\text{Var}(\mathbf{e}_{n|q}) = \sum_{i=0}^d \sigma_{(n+1,i)}^2 - \sum_{i=0}^d \sigma_{(n,i)}^2 \\
&=\sum_{i=0}^d a_{i}^2 \sigma_{(n,i)}^2 + \sum_{i=0}^d\hat{\sigma}_{i}^2 - \sum_{i=0}^d \sigma_{(n,i)}^2 \\
&=\sum_{i=0}^d a_{i}^2 [a_{i}^2 \sigma_{(n-1,i)}^2 + \hat{\sigma}_{i}^2] - \sum_{i=0}^d a_{i}^2 \sigma_{(n-1,i)}^2\\
&=\sum_{i=0}^d a_{i}^2 [(a_{i}^2 - 1) \sigma_{(n-1,i)}^2 + \hat{\sigma}_{i}^2] 
\end{align}
Using our inductive assumption in Equation \eqref{euqa:var3} and $a_{i}^2 \geqslant 0$,
it shows that $\text{Var}(\mathbf{e}_{n+1|q})-\text{Var}(\mathbf{e}_{n|q}) \geqslant 0$ is true for $k = n + 1$.
By mathematical induction, Equation \eqref{euqa:var2} is proven to be true for all natural numbers $n$.

As a result, $H(\mathbf{e}_{k|q}) \geqslant H(\mathbf{e}_{k-1|q})$ holds in three types of affine transformations utilized in this paper, the theorem is proved.

\end{proof}

\subsection{Proof for Theorem \ref{theo:addpath}} 
\label{app:t2}

{\em 
Let $\mathcal{P}$ be the set of all shortest paths between the query entity $q$ and a candidate entity $e_t$ in $\mathcal{G}$. If a redundant path $\hat{P}$ is added to the path set $\mathcal{P}$, then the transformation error entropy of the mean-aggregated vector $\mathbf{e}_{t|q}$ would increase, i.e. $H(\mathcal{F}(\mathcal{P} \cup {\hat{P}})) > H(\mathcal{F}(\mathcal{P}))$.
}

\begin{proof}

Given the path set $\mathcal{P}=\{P_1,P_2,\cdots,P_m\}$ in which $m$ is the number of the shortest paths, we can compute the transformation error variance of $\mathcal{F}(\mathcal{P})$ as follows:
\begin{align}
\text{Var}(\mathcal{F}(\mathcal{P})) = \frac{1}{m^2} \big( \sum_{i=1}^{m} \text{Var}(\mathbf{P}_i) +  \sum_{i \neq j}\text{Cov}(\epsilon_{i},\epsilon_{j}) \big),
\end{align}
where $\mathbf{P}_i, \epsilon_{i}$ denote the embedding vector and the transformation error of the path $P_i$, and \text{Cov}() is the covariance of two error variables. To simplify the following derivation, we adjust the single-path error variance in Equation \ref{euqa:var0}:
\begin{align}
\text{Var}(\mathbf{P}_i) = \text{Var}(\mathbf{e}_{\ell|q}) = \text{Var}(\mathbf{e}_{\ell-1|q}) + \hat{\sigma}^2  = \text{Var}(\mathbf{q}) + \ell\hat{\sigma}^2 = \ell\hat{\sigma}^2,
\end{align}
which means the variance of input errors won’t be changed (neither enlarged nor shrunk) by embedding or GNN operations. Such that, given the path $P_i$ from $q$ containing $\ell$ triples, the error variance of the path $\sigma_{P_i}^2 = \ell\hat{\sigma}^2 > 0$. 

Then, we can represent the aggregated variance $\text{Var}(\mathcal{F}(\mathcal{P}))$ as:
\begin{align}
\text{Var}(\mathcal{F}(\mathcal{P})) = \frac{1}{m^2} \big( \sum_{i=1}^{m} \sigma_{P_i}^2 +  \sum_{i \neq j}\text{Cov}(\epsilon_{i},\epsilon_{j}) \big) = \frac{1}{m^2} \sum_{i=1}^{m} k_i\hat{\sigma}^2 +  \frac{C_m}{m^2}  = \frac{\ell}{m} \hat{\sigma}^2 +  \frac{C_m}{m^2} ,
\end{align}
where $C_m = \sum_{i \neq j}\text{Cov}(\epsilon_{i},\epsilon_{j})$ denotes the covariance value derived from the path overlapping.

Now, we discuss the changes in the error variance after adding a new path $\hat{P}$. Because $\hat{P}$ contains at least $\ell$ triples that exist in $\mathcal{P}$, without loss of generality, we first assume that $\hat{P}$ internally contains an independent part with length $\alpha>0$, and the remainder of $\hat{P}$ with length $\ell$ is exactly an existing path $p_l \in \mathcal{P}$. Then, the changed error variance can be computed as follows:
\begin{align}
\label{euqa:t2addp}
&\text{Var}(\mathcal{F}(\mathcal{P}\cup \{\hat{P}\})) = \frac{1}{(m+1)^2} \big( \sum_{i=1}^{m} \sigma_{P_i}^2 +  \sum_{i \neq j}\text{Cov}(\epsilon_{i},\epsilon_{j}) + \sigma_{\hat{P}}^2 + 2 \sum_{i=1}^{m}\text{Cov}(\epsilon_{i},\epsilon_{l}) \big)\\
\label{euqa:t2approx}
&\approx \frac{1}{(m+1)^2} \big( \sum_{i=1}^{m} \sigma_{P_i}^2 +  C_m + \sigma_{\hat{p}}^2 + 2(\text{Cov}(\epsilon_{\hat{P}},\epsilon_{l}) + \frac{1}{m}C_m)\big) \\
&= \frac{1}{(m+1)^2} \big( m\ell \hat{\sigma}^2 +  C_m + (\ell+\alpha)\hat{\sigma}^2  + 2\ell\hat{\sigma}^2 + \frac{2}{m}C_m \big)\\
&= \frac{m\ell +3\ell + \alpha}{(m+1)^2} \hat{\sigma}^2 + \frac{m+2}{m(m+1)^2} C_m,
\end{align}

Note that, in Equation \ref{euqa:t2approx}, due to the path overlapping in $\mathcal{P}$ is agnostic, we assume the covariance of one path is the average value of the total covariance $C_m$, i.e. $\sum_{i\neq l}\text{Cov}(\epsilon_{i},\epsilon_{l}) \approx \frac{1}{m}C_m$. 

Then, we need to prove $H(\mathcal{F}(\mathcal{P} \cup {\hat{P}})) > H(\mathcal{F}(\mathcal{P}))$, and we simplify the equation as follows:
\begin{align}
& H(\mathcal{F}(\mathcal{P} \cup {\hat{P}})) - H(\mathcal{F}(\mathcal{P}))=\frac{m\ell +3\ell + \alpha}{(m+1)^2} \hat{\sigma}^2 + \frac{m+2}{m(m+1)^2} C_m - \big( \frac{\ell}{m} \hat{\sigma}^2 + \frac{C_m}{m^2} \big)\\
& =\frac{m^2(m\ell +3\ell + \alpha)\hat{\sigma}^2 + m(m+2)C_m - m(m+1)^2\ell\hat{\sigma}^2 - (m+1)^2 C_m}{m^2(m+1)^2} \\
& =\frac{(m(m-1)\ell + m^2 \alpha)\hat{\sigma}^2 - C_m}{m^2(m+1)^2}.
\label{euqa:t2ine1}
\end{align}

Meanwhile, we know that the total covariance $C_m$ reaches its maximum value when all $m$ paths fully overlap:
\begin{align}
C_m = \sum_{i \neq j}\text{Cov}(\epsilon_{i},\epsilon_{j}) \big) \leqslant \sum_{i \neq j}\ell\hat{\sigma}^2 = m(m-1)\ell\hat{\sigma}^2.
\label{euqa:t2ine2}
\end{align}

Combining Equation \ref{euqa:t2ine1} and Equation \ref{euqa:t2ine2}, we have:
\begin{align}
&H(\mathcal{F}(\mathcal{P} \cup {\hat{P}})) - H(\mathcal{F}(\mathcal{P})) = \frac{m^2\alpha\hat{\sigma}^2 + m(m-1)\ell\hat{\sigma}^2 - C_m}{m^2(m+1)^2} \geqslant \frac{\alpha\hat{\sigma}^2}{(m+1)^2} > 0.
\end{align}

Therefore, $H(\mathcal{F}(\mathcal{P} \cup {\hat{P}})) > H(\mathcal{F}(\mathcal{P}))$ holds, when the new path $\hat{P}$ can be divided into two parts, an independent part with length $\alpha>0$ and an existing path $p_l$. 

Then, considering the general form of $\hat{P}$ having $\ell$ triples in $\mathcal{P}$, we need to recompute the covariance term in Equation \ref{euqa:t2addp}. Let $\hat{P}_\ell$ denotes the $\ell$ triples and $\hat{P}_\alpha$ is the additional part with length $\alpha$. Due to the assumption that the transformation errors in all triples are i.i.d, the covariance of $\hat{P}_\ell$ is equal to that of the path $P_l$. Meanwhile, if the path part $\hat{P}_\alpha$ is overlapped with some paths in $\mathcal{P}$, the total covariance would further increase. Such that, we have:
\begin{align}
\sum_{i=1}^{m}\text{Cov}(\epsilon_{i},\epsilon_{\hat{P}}) = (\sum_{i=1}^{m}\text{Cov}(\epsilon_{i},\epsilon_{\hat{P}_\alpha}) + \sum_{i=1}^{m}\text{Cov}(\epsilon_{i},\epsilon_{\hat{P}_\ell})) \geqslant \sum_{i=1}^{m}\text{Cov}(\epsilon_{i},\epsilon_{l}),
\end{align}

Therefore, $H(\mathcal{F}(\mathcal{P} \cup {\hat{P}})) > H(\mathcal{F}(\mathcal{P}))$ holds, when adding a new path $\hat{P}$ having $\ell$ triples in $\mathcal{P}$. The theorem is proved.

\end{proof}

\subsection{Proof for Proposition \ref{theo:addedge}} 
\label{app:t3}

{\em 
Let $\mathcal{\hat{G}}$ be the KG subgraph constructed by the triples of all shortest paths starting from the query entity $q$. Suppose a new triple $t = (e_1, r, e_2)$ whose head and tail entities exist in $\mathcal{\hat{G}}$ and the relative distance $\gamma_{e_1} \geqslant \gamma_{e_2}$. Adding this triple $t$ into $\mathcal{\hat{G}}$ leads to at least one new redundant path passing $t$. 
}

\begin{proof}
Since the entity $e_1$ exists in $\mathcal{\hat{G}}$, there must be one shortest path $P$ to a candidate entity $e_t$ passing $e_1$.

If the path $P$ also passes $e_2$, we can divide $P$ into three parts: $P_{q,e_2}$, $P_{e_2,e_1}$, and $P_{e_1,e_t}$. It is easy to prove that the path sequentially connecting $\{P_{q,e_2}, P_{e_2,e_1}, (e_1, r, e_2), P_{e_2,e_1}, P_{e_1,e_t}\}$ is a new redundant path.

If the path $P$ does not pass $e_2$, there must be another shortest path $P'$ to a candidate entity $e_{t'}$ via $e_2$. In this case, we can divide $P$ and $P'$ into four parts: $P_{q,e_1}$, $P_{e_1,e_t}$, $P'_{q,e_2}$, and $P_{e_2,e_{t'}}$. There is a new path sequentially connecting $\{P_{q,e_1}, (e_1, r, e_2), P_{e_2,e_{t'}}\}$. Because the relative distance of $e_1$ is not smaller than that of $e_2$, the length of $P_{q,e_1}$ is not shorter than that of $P'_{q,e_2}$. Such that, the length of the new path is longer than the shortest path. It is a new redundant path.

Overall, a new redundant path exists after adding such a new triple, so the theorem is proved.

\end{proof}

\begin{table}[!ht]
    \centering
    \caption{Hyperparameter configurations of GraPE on different datasets.}
    \label{tab:hyperparameter}
    \begin{adjustbox}{max width=\textwidth}
        \begin{tabular}{llccccc}
    \toprule
    \multicolumn{2}{l}{\bf{Task Setting}}
       & \multicolumn{2}{c}{\bf{Transductive Reasoning}} & \multicolumn{3}{c}{\bf{Inductive Reasoning}} \\
    \multicolumn{2}{l}{\bf{Hyperparameter}}
       & \bf{FB15k-237} & \bf{WN18RR} & \bf{FB15k-237} & \bf{WN18RR} & \bf{NELL-995} \\
    \midrule
    \multirow{3}{*}{\bf{Architecture}}
       & layer(L) & 3         & 5      & 4    & 5        & 5      \\
       & encoder dim ($d$).  & 64        & 64     & 32   & 32       & 32     \\
       & decoder dim ($d_l$).  & 8        & 8     & 8   & 8       & 8     \\
    \midrule
    \multirow{2}{*}{\bf{Function}}
       & transform     & DistMult         & DistMult      & DistMult    & DistMult        & DistMult      \\
       & aggregate & PNA        & PNA     & PNA   & PNA       & PNA    \\
    \midrule
    \multirow{4}{*}{\bf{Learning}}
       & optimizer     & Adam   & Adam   & Adam   & Adam   & Adam    \\
       & batch size    & 16     & 16     & 16     & 16     & 16      \\
       & learning rate & 5e-3   & 5e-3   & 5e-4   & 5e-4   & 5e-4    \\
       & epoch       & 20     & 20     & 20     & 20     & 20      \\       
    \bottomrule
        \end{tabular}
    \end{adjustbox}
\end{table}

\begin{table}[!ht]
    \centering
    \caption{Statistics of transductive benchmark datasets.}
    \label{table:ds-tra}
    \begin{adjustbox}{max width=\textwidth}
            \begin{tabular}{lccccc}
    \toprule
    \multirow{2}{*}{Dataset} & \multirow{2}{*}{$|\mathcal{E}|$} & \multirow{2}{*}{$|\mathcal{R}|$} & \multicolumn{3}{c}{$|\mathcal{F}|$} \\
    & & & {\#Train} & {\#Validation} & {\#Test} \\
    \midrule
    FB15k-237~\cite{FB15k237} & 14,541 & 237 & 272,115 & 17,535 & 20,466 \\
    WN18RR~\cite{WN18RR} & 40,943 & 11 & 86,835 & 3,034 & 3,134 \\
    \bottomrule
            \end{tabular}
    \end{adjustbox}
\end{table}

\begin{table}[tp]
    \caption{Statistics of inductive benchmark datasets.}
    \label{table:ds-ind}
    \centering
    \small
    \begin{tabular}{@{}p{3pt}p{12pt}p{5pt}p{14pt}lp{5pt}p{14pt}lp{5pt}p{14pt}l@{}}
        \toprule
        & & \multicolumn{3}{c}{WN18RR} & \multicolumn{3}{c}{FB15k-237} & \multicolumn{3}{c}{NELL-995} \\
        & & $|\mathcal{R}|$ & $|\mathcal{E}|$ & $|\mathcal{F}|$ & $|\mathcal{R}|$ & $|\mathcal{E}|$ & $|\mathcal{F}|$ & $|\mathcal{R}|$ & $|\mathcal{E}|$ & $|\mathcal{F}|$ \\
        \midrule
        \multirow{2}{*}{v1} & train & 9 & 2,746 & 6,678 & 183 & 2,000 & 5,226 & 14 &10,915 & 5,540 \\
        & \textit{test} & 9 & 922 & 1,991 & 146 & 1,500 & 2,404 & 14 & 225 & 1,034 \\
        \midrule
        \multirow{2}{*}{v2} & train & 10 & 6,954 & 18,968 & 203 & 3,000 & 12,085 & 88 & 2,564 & 10,109 \\
        & \textit{test} & 10 & 2,923 & 4,863 & 176 & 2,000 & 5,092 & 79 & 4,937 & 5,521 \\
        \midrule
        \multirow{2}{*}{v3} & train & 11 & 12,078 & 32,150 & 218 & 4,000 & 22,394 & 142 & 4,647 & 20,117 \\
        & \textit{test} & 11 & 5,084 & 7,470 & 187 & 3,000 & 9,137 & 122 & 4,921 & 9,668 \\
        \midrule
        \multirow{2}{*}{v4} & train & 9 & 3,861 & 9,842 & 222 & 5,000 & 33,916 & 77 & 2,092 & 9289 \\
        & \textit{test} & 9 & 7,208 & 15,157 & 204 & 3,500 & 14,554 & 61 & 3,294 & 8,520 \\
        \bottomrule
    \end{tabular}
\end{table}

\section{Model Details and Implementation} 
\label{app:md}

GraPE is expected to absorb effective components in both KGE and GNN areas.
To prove the feasibility of the entropy-guided percolation and the graph percolation process, we set the input query vector $\mathbf{q}$ as the all-ones vector. And each relative embedding vector $\mathbf{e}_{t|q}$ is initialized as the zero vector. It will be our future work to explore more effective initialization settings.

Then, we discuss the design space of the detailed functions in the GNN layer, including the transform operator $\otimes$, the aggregation function $\phi(\cdot)$, and the relation parameters $\mathbf{r}$.

For the transform operator $\otimes$, traditional KGE models provide multiple selections, such as vector addition in TransE \cite{TransE}, vector multiplication in Distmult \cite{DistMult} and vector rotation in RotatE \cite{RotatE}. Although there are more complicated scoring functions \cite{ConvE,GoogleAttH,OurRotL}, we instantiate the above three efficient operators in GraPE. Compared with typical KGE models using static scoring functions, GraPE has a stronger representational capability due to the neural networks in GNN layers. 

The aggregation function $\phi(\cdot)$ is the key component of the GNN layer, and we specify it in GraPE to be SUM, MEAN, and PNA\cite{PNA-NIPS20}. The original PNA aggregator jointly utilizes four types of aggregations which is computationally intensive. We simplify it by only using MEAN and STD aggregations. It is worth noting that we utilize the global degree value of each entity for averaging in MEAN and PNA, to avoid the problem of indistinguishable entities on local subgraphs.

The relation parameters $\mathbf{r} \in \mathbb{R}^d$ are the major trainable parameters in GraPE. In order to capture the dependencies between the triple relation and the query relation $r_q$, we follow NBFNet \cite{NBF-NIPS21} to generate relation embeddings $\{\mathbf{r}_{i|q}\}$ via a linear function over the query relation but only use $O\big(|\mathcal{R}|d+d^2\big)$ parameters. Furthermore, for two GNN layers in GraPE, we utilize different dimensions of relation parameters, i.e. $d$ and $d_l$.

About the model hyperparameters, we list the default hyperparameter configurations of GraPE on different datasets in Table \ref{tab:hyperparameter}.
Note for inductive settings, we use the same hyperparameters for four sub-datasets. 
All the hyperparameters are chosen by the performance on the validation set.

\section{Model Computational Complexity} 
\label{app:cc}

\textbf{Space Complexity:} 
GraPE outperforms previous KG reasoning methods in terms of space complexity.
Traditional KGE models using absolute entity embeddings require parameter storage costs as $O\big( |\mathcal{E}|d + |\mathcal{R}|d\big)$, 
GraPE and GNN-based methods using relative knowledge embeddings, such as GraIL \cite{GraIL-ICML19}, NBFNet \cite{NBF-NIPS21} and RED-GNN \cite{REDGNN-WWW22} have a lower complexity around $O\big(|\mathcal{R}|d + d^2\big)$, because $|\mathcal{E}| \gg |\mathcal{R}|$ in large-scale KGs. Furthermore, denote the parameter amount in each GNN layer as $d\Theta_R$, GNN-based methods require at least $O\big(Ld\Theta_R)$ parameters. GraPE with two GNN layers contains significantly fewer parameters whose amount is $O\big(d\Theta_R + d^l\Theta_R)$.

\textbf{Time Complexity:} 
GraPE has a lower time complexity in the inference period compared with the recent three GNN-based methods.
Considering the inference process with one query, its time complexity $O\big(|\mathcal{\hat{T}}|d+|\mathcal{E}|d)$ of these GNN-based models is mainly determined by the calculated triple amount $|\mathcal{\hat{T}}|$.
Therefore, we compare these GNN-based methods by quantifying the calculation times of involved triples for a query. Given the standard $L$-hop neighborhood subgraph $\mathcal{G}_{q}$ of the entity $q$, the $\ell$-th hop contains triples as $\{(e_h,r,e_t)|e_h,e_t \in \mathcal{N}_{q}^{\ell-1}\cup\mathcal{N}_{q}^{\ell}\}$ and we denote its amount as $n_\ell$. Then, the total triple amount in $\mathcal{G}_{q}$ is denoted as $N_{q}^L = \sum_{\ell=1}^L{n_\ell}$. A detailed comparison is as follows:

\begin{itemize}[leftmargin=*]
\item \textbf{GraIL \cite{GraIL-ICML19}} extracts an enclosing subgraph from $\mathcal{G}_{e_q}$ and conducts the $L$-layer GNN propagation for each candidate entity $e_t$. So the calculated triple amount is $N_{GraIL}=L|\mathcal{E}|N_{q, e_t}^L > LN_{q}^L$.
\item \textbf{NBFNet \cite{NBF-NIPS21}} calculates once $L$-layer GNN propagation for a query on the whole graph. We can represent its calculated triple amount as $N_{NBFNet}=L(min(|\mathcal{T}'|, N_{q}^L)) < N_{GraIL}$.
\item \textbf{RED-GNN \cite{REDGNN-WWW22}} only calculates involved triples from the first hop to the $\ell$-th hop in the $\ell$-th GNN layer. Its calculated triples are fewer than the global propagation in NBFNet, i.e. $N_{RED-GNN}=N_{q}^L + \sum_{\ell=1}^{L-1} {\sum_{i=1}^\ell{n_i}} < N_{NBFNet}$.
\item \textbf{GraPE} contains a $L$-$1$ layer graph percolation process and a 1-layer GNN propagation on the whole subgraph. Considering its calculated triples in each $\ell$-th percolation layer are fewer than $n_\ell$, we have $N_{GraPE} < N_{q}^L + \sum_{\ell=1}^{L-1}{n_\ell} < N_{RED-GNN}$.
\end{itemize}

\textbf{Training Complexity:}
The process of subgraph indexing and mini-batch training in GraPE is following the RED-GNN approach \cite{REDGNN-WWW22}. The $\ell$-hop neighbors $N_{q}^\ell$ are constructed by the 1-hop neighbors of entities in the $N_{q}^{\ell-1}$. And we record the out-going triple ids of every entity in the form of a sparse matrix, whose reading complexity for one entity is $O\big(1)$ and the space complexity is $O\big(|\hat{\mathcal{T}}|)$. Given the $m$ entities in $N_{q}^{\ell-1}$ and the average triple number $D$, we can collect the 1-hop neighbors and triples with the time complexity $O\big(mD)$.

As shown in Algorithm \ref{alg:gip}, for one query, we progressively load 1-hop neighbors and triples from the sparse matrix, and then index them for each GNN iteration. The space and time complexity of related operations can be linear with the number of entities or triples in each layer (The entity/triple amount in each layer is usually much smaller than the whole KG). Besides, GraPE is parallelizable and supports mini-batch training.  Following RED-GNN, in each iteration, we reindex all the nodes in different query subgraphs and construct a whole subgraph (two same entities in different subgraphs will be distinguished). Therefore, given the batch size $B$, the space and time cost of batch operations for subgraph indexing is approximately $B$ times that of one query.

%% file: main.bbl
\begin{thebibliography}{42}
\providecommand{\natexlab}[1]{#1}
\providecommand{\url}[1]{\texttt{#1}}
\expandafter\ifx\csname urlstyle\endcsname\relax
  \providecommand{\doi}[1]{doi: #1}\else
  \providecommand{\doi}{doi: \begingroup \urlstyle{rm}\Url}\fi

\bibitem[Bordes et~al.(2013)Bordes, Garc{\'{\i}}a{-}Dur{\'{a}}n, Weston, and
  Yakhnenko]{TransE}
Antoine Bordes, Alberto Garc{\'{\i}}a{-}Dur{\'{a}}n, Jason Weston, and Oksana
  Yakhnenko.
\newblock Translating embeddings for modeling multi-relational data.
\newblock In \emph{Proceedings of the 27th Annual Conference on Neural
  Information Processing Systems, December 5-8, 2013, Lake Tahoe, Nevada,
  United States}, pages 2787--2795, 2013.

\bibitem[Bordes et~al.(2014)Bordes, Glorot, Weston, and Bengio]{WN18RR}
Antoine Bordes, Xavier Glorot, Jason Weston, and Yoshua Bengio.
\newblock {A Semantic Matching Energy Function for Learning with
  Multi-relational Data}.
\newblock \emph{Machine Learning}, 94:\penalty0 233--259, 2014.

\bibitem[Chami et~al.(2020)Chami, Wolf, Juan, Sala, Ravi, and
  R{\'e}]{GoogleAttH}
Ines Chami, Adva Wolf, Da-Cheng Juan, Frederic Sala, Sujith Ravi, and
  Christopher R{\'e}.
\newblock Low-dimensional hyperbolic knowledge graph embeddings.
\newblock In \emph{Proceedings of the 58th Annual Meeting of the Association
  for Computational Linguistics, {ACL} 2020, Online, July 5-10, 2020}, pages
  6901--6914, 2020.

\bibitem[Chan et~al.(2021)Chan, Xu, Long, Sanyal, Gupta, and Ren]{NIPS21-KG1}
Aaron Chan, Jiashu Xu, Boyuan Long, Soumya Sanyal, Tanishq Gupta, and Xiang
  Ren.
\newblock Salkg: Learning from knowledge graph explanations for commonsense
  reasoning.
\newblock In \emph{Advances in Neural Information Processing Systems 34: Annual
  Conference on Neural Information Processing Systems 2021, NeurIPS 2021,
  December 6-14, 2021, virtual}, pages 18241--18255, 2021.

\bibitem[Corso et~al.(2020)Corso, Cavalleri, Beaini, Li{\`{o}}, and
  Velickovic]{PNA-NIPS20}
Gabriele Corso, Luca Cavalleri, Dominique Beaini, Pietro Li{\`{o}}, and Petar
  Velickovic.
\newblock Principal neighbourhood aggregation for graph nets.
\newblock In \emph{Advances in Neural Information Processing Systems 33: Annual
  Conference on Neural Information Processing Systems 2020, NeurIPS 2020,
  December 6-12, 2020, virtual}, 2020.

\bibitem[Darcy(1856)]{darcy1856}
Henry Darcy.
\newblock \emph{Les fontaines publiques de la ville de Dijon}, volume~1.
\newblock Victor Dalmont, Paris, 1856.

\bibitem[Das et~al.(2017{\natexlab{a}})Das, Dhuliawala, Zaheer, Vilnis,
  Durugkar, Krishnamurthy, Smola, and McCallum]{MINERVA-NIPS17}
Rajarshi Das, Shehzaad Dhuliawala, Manzil Zaheer, Luke Vilnis, Ishan Durugkar,
  Akshay Krishnamurthy, Alex Smola, and Andrew McCallum.
\newblock Go for a walk and arrive at the answer: Reasoning over knowledge
  bases with reinforcement learning.
\newblock In \emph{6th Workshop on Automated Knowledge Base Construction,
  AKBC@NIPS 2017, Long Beach, California, USA, December 8, 2017}.
  OpenReview.net, 2017{\natexlab{a}}.

\bibitem[Das et~al.(2017{\natexlab{b}})Das, Neelakantan, Belanger, and
  McCallum]{RNNChain-EACL17}
Rajarshi Das, Arvind Neelakantan, David Belanger, and Andrew McCallum.
\newblock Chains of reasoning over entities, relations, and text using
  recurrent neural networks.
\newblock In \emph{Proceedings of the 15th Conference of the European Chapter
  of the Association for Computational Linguistics, {EACL} 2017, Valencia,
  Spain, April 3-7, 2017, Volume 1: Long Papers}, pages 132--141. Association
  for Computational Linguistics, 2017{\natexlab{b}}.

\bibitem[Dettmers et~al.(2018)Dettmers, Minervini, Stenetorp, and
  Riedel]{ConvE}
Tim Dettmers, Pasquale Minervini, Pontus Stenetorp, and Sebastian Riedel.
\newblock Convolutional 2d knowledge graph embeddings.
\newblock In \emph{Proceedings of the Thirty-Second {AAAI} Conference on
  Artificial Intelligence, (AAAI-18), New Orleans, Louisiana, USA, February
  2-7, 2018}, pages 1811--1818, 2018.

\bibitem[Elger et~al.(2020)Elger, LeBret, Crowe, and Roberson]{pbook3}
Donald~F Elger, Barbara~A LeBret, Clayton~T Crowe, and John~A Roberson.
\newblock \emph{Engineering fluid mechanics}.
\newblock John Wiley \& Sons, 2020.

\bibitem[Finnemore and Franzini(2002)]{pbook2}
E~John Finnemore and Joseph~B Franzini.
\newblock \emph{Fluid mechanics with engineering applications}.
\newblock McGraw-Hill Education, 2002.

\bibitem[Ji et~al.(2022)Ji, Pan, Cambria, Marttinen, and Yu]{2020survey}
Shaoxiong Ji, Shirui Pan, Erik Cambria, Pekka Marttinen, and Philip~S. Yu.
\newblock A survey on knowledge graphs: Representation, acquisition, and
  applications.
\newblock \emph{{IEEE} Trans. Neural Networks Learn. Syst.}, 33\penalty0
  (2):\penalty0 494--514, 2022.

\bibitem[Lin et~al.(2015)Lin, Liu, Luan, Sun, Rao, and Liu]{PTransE}
Yankai Lin, Zhiyuan Liu, Huan{-}Bo Luan, Maosong Sun, Siwei Rao, and Song Liu.
\newblock Modeling relation paths for representation learning of knowledge
  bases.
\newblock In \emph{Proceedings of the 2015 Conference on Empirical Methods in
  Natural Language Processing, {EMNLP} 2015, Lisbon, Portugal, September 17-21,
  2015}, pages 705--714. The Association for Computational Linguistics, 2015.

\bibitem[Liu et~al.(2021)Liu, Grau, Horrocks, and Kostylev]{INDIGO-NIPS21}
Shuwen Liu, Bernardo~Cuenca Grau, Ian Horrocks, and Egor~V. Kostylev.
\newblock {INDIGO:} gnn-based inductive knowledge graph completion using
  pair-wise encoding.
\newblock In Marc'Aurelio Ranzato, Alina Beygelzimer, Yann~N. Dauphin, Percy
  Liang, and Jennifer~Wortman Vaughan, editors, \emph{Advances in Neural
  Information Processing Systems 34: Annual Conference on Neural Information
  Processing Systems 2021, NeurIPS 2021, December 6-14, 2021, virtual}, pages
  2034--2045, 2021.

\bibitem[Meilicke et~al.(2018)Meilicke, Fink, Wang, Ruffinelli, Gemulla, and
  Stuckenschmidt]{RuleN-2018}
Christian Meilicke, Manuel Fink, Yanjie Wang, Daniel Ruffinelli, Rainer
  Gemulla, and Heiner Stuckenschmidt.
\newblock Fine-grained evaluation of rule- and embedding-based systems for
  knowledge graph completion.
\newblock In \emph{The Semantic Web - {ISWC} 2018 - 17th International Semantic
  Web Conference, Monterey, CA, USA, October 8-12, 2018, Proceedings, Part
  {I}}, volume 11136 of \emph{Lecture Notes in Computer Science}, pages 3--20.
  Springer, 2018.

\bibitem[Neelakantan et~al.(2015)Neelakantan, Roth, and McCallum]{ComVSM-ACL15}
Arvind Neelakantan, Benjamin Roth, and Andrew McCallum.
\newblock Compositional vector space models for knowledge base completion.
\newblock In \emph{Proceedings of the 53rd Annual Meeting of the Association
  for Computational Linguistics and the 7th International Joint Conference on
  Natural Language Processing of the Asian Federation of Natural Language
  Processing, {ACL} 2015, July 26-31, 2015, Beijing, China, Volume 1: Long
  Papers}, pages 156--166. The Association for Computer Linguistics, 2015.

\bibitem[Pan et~al.(2022)Pan, Ye, Han, Song, and Huang]{NIPS22-KG3}
Xuran Pan, Tianzhu Ye, Dongchen Han, Shiji Song, and Gao Huang.
\newblock Contrastive language-image pre-training with knowledge graphs.
\newblock In \emph{Advances in Neural Information Processing Systems}, 2022.

\bibitem[Qu et~al.(2021)Qu, Chen, Xhonneux, Bengio, and Tang]{RNNLogic-ICLR21}
Meng Qu, Junkun Chen, Louis{-}Pascal A.~C. Xhonneux, Yoshua Bengio, and Jian
  Tang.
\newblock Rnnlogic: Learning logic rules for reasoning on knowledge graphs.
\newblock In \emph{9th International Conference on Learning Representations,
  {ICLR} 2021, Virtual Event, Austria, May 3-7, 2021}. OpenReview.net, 2021.

\bibitem[Sadeghian et~al.(2019)Sadeghian, Armandpour, Ding, and
  Wang]{DRUM-NIPS19}
Ali Sadeghian, Mohammadreza Armandpour, Patrick Ding, and Daisy~Zhe Wang.
\newblock {DRUM:} end-to-end differentiable rule mining on knowledge graphs.
\newblock In \emph{Advances in Neural Information Processing Systems 32: Annual
  Conference on Neural Information Processing Systems 2019, NeurIPS 2019,
  December 8-14, 2019, Vancouver, BC, Canada}, pages 15321--15331, 2019.

\bibitem[Schetz and Fuhs(1999)]{pbook1}
Joseph~A Schetz and Allen~E Fuhs.
\newblock \emph{Fundamentals of fluid mechanics}.
\newblock John Wiley \& Sons, 1999.

\bibitem[Schlichtkrull et~al.(2018)Schlichtkrull, Kipf, Bloem, van~den Berg,
  Titov, and Welling]{RGCN}
Michael~Sejr Schlichtkrull, Thomas~N. Kipf, Peter Bloem, Rianne van~den Berg,
  Ivan Titov, and Max Welling.
\newblock Modeling relational data with graph convolutional networks.
\newblock In \emph{Proceedings of the 15th Extended Semantic Web Conference,
  {ESWC} 2018, Heraklion, Crete, Greece, June 3-7, 2018}, pages 593--607, 2018.

\bibitem[Sun et~al.(2019)Sun, Deng, Nie, and Tang]{RotatE}
Zhiqing Sun, Zhi{-}Hong Deng, Jian{-}Yun Nie, and Jian Tang.
\newblock Rotate: Knowledge graph embedding by relational rotation in complex
  space.
\newblock In \emph{Proceedings of the 7th International Conference on Learning
  Representations, {ICLR} 2019, New Orleans, LA, USA, May 6-9, 2019}, 2019.

\bibitem[Teru et~al.(2020)Teru, Denis, and Hamilton]{GraIL-ICML19}
Komal~K. Teru, Etienne~G. Denis, and William~L. Hamilton.
\newblock Inductive relation prediction by subgraph reasoning.
\newblock In \emph{Proceedings of the 37th International Conference on Machine
  Learning, {ICML} 2020, 13-18 July 2020, Virtual Event}, volume 119 of
  \emph{Proceedings of Machine Learning Research}, pages 9448--9457. {PMLR},
  2020.

\bibitem[Toutanova and Chen(2015)]{FB15k237}
Kristina Toutanova and Danqi Chen.
\newblock Observed versus latent features for knowledge base and text
  inference.
\newblock In \emph{Proceedings of the 3rd Workshop on Continuous Vector Space
  Models and their Compositionality}, pages 57--66, 2015.

\bibitem[Vashishth et~al.(2020)Vashishth, Sanyal, Nitin, and
  Talukdar]{CompGCN-ICLR20}
Shikhar Vashishth, Soumya Sanyal, Vikram Nitin, and Partha~P. Talukdar.
\newblock Composition-based multi-relational graph convolutional networks.
\newblock In \emph{8th International Conference on Learning Representations,
  {ICLR} 2020, Addis Ababa, Ethiopia, April 26-30, 2020}. OpenReview.net, 2020.

\bibitem[Wang et~al.(2021{\natexlab{a}})Wang, Ren, and Leskovec]{PathCon-KDD21}
Hongwei Wang, Hongyu Ren, and Jure Leskovec.
\newblock Relational message passing for knowledge graph completion.
\newblock In \emph{{KDD} '21: The 27th {ACM} {SIGKDD} Conference on Knowledge
  Discovery and Data Mining, Virtual Event, Singapore, August 14-18, 2021},
  pages 1697--1707. {ACM}, 2021{\natexlab{a}}.

\bibitem[Wang et~al.(2020)Wang, Liu, Xu, and Sheng]{OurComp}
Kai Wang, Yu~Liu, Xiujuan Xu, and Quan~Z. Sheng.
\newblock Enhancing knowledge graph embedding by composite neighbors for link
  prediction.
\newblock \emph{Computing}, 102\penalty0 (12):\penalty0 2587--2606, 2020.

\bibitem[Wang et~al.(2021{\natexlab{b}})Wang, Liu, Lin, and Sheng]{OurRotL}
Kai Wang, Yu~Liu, Dan Lin, and Michael Sheng.
\newblock Hyperbolic geometry is not necessary: Lightweight euclidean-based
  models for low-dimensional knowledge graph embeddings.
\newblock In \emph{Findings of the Association for Computational Linguistics:
  {EMNLP} 2021, Virtual Event / Punta Cana, Dominican Republic, 16-20 November,
  2021}, pages 464--474, 2021{\natexlab{b}}.

\bibitem[Wang et~al.(2021{\natexlab{c}})Wang, Liu, Ma, and Sheng]{OurMulDE}
Kai Wang, Yu~Liu, Qian Ma, and Quan~Z. Sheng.
\newblock Mulde: Multi-teacher knowledge distillation for low-dimensional
  knowledge graph embeddings.
\newblock In \emph{Proceedings of the {WWW} '21: The Web Conference 2021,
  Virtual Event / Ljubljana, Slovenia, April 19-23, 2021}, pages 1716--1726,
  2021{\natexlab{c}}.

\bibitem[Wang et~al.(2022{\natexlab{a}})Wang, Liu, and Sheng]{OurHaLE}
Kai Wang, Yu~Liu, and Quan~Z. Sheng.
\newblock Swift and sure: Hardness-aware contrastive learning for
  low-dimensional knowledge graph embeddings.
\newblock In \emph{{WWW} '22: The {ACM} Web Conference 2022, Virtual Event,
  Lyon, France, April 25 - 29, 2022}, pages 838--849. {ACM},
  2022{\natexlab{a}}.

\bibitem[Wang et~al.(2022{\natexlab{b}})Wang, Li, Sun, Liu, Li, Yin, and
  Abdelzaher]{NIPS22-KG2}
Ruijie Wang, Zheng Li, Dachun Sun, Shengzhong Liu, Jinning Li, Bing Yin, and
  Tarek~F. Abdelzaher.
\newblock Learning to sample and aggregate: Few-shot reasoning over temporal
  knowledge graphs.
\newblock In \emph{Advances in Neural Information Processing Systems},
  2022{\natexlab{b}}.

\bibitem[Xiong et~al.(2017)Xiong, Hoang, and Wang]{NELL995}
Wenhan Xiong, Thien Hoang, and William~Yang Wang.
\newblock Deeppath: {A} reinforcement learning method for knowledge graph
  reasoning.
\newblock In \emph{Proceedings of the 2017 Conference on Empirical Methods in
  Natural Language Processing, {EMNLP} 2017, Copenhagen, Denmark, September
  9-11, 2017}, pages 564--573. Association for Computational Linguistics, 2017.

\bibitem[Yang et~al.(2015)Yang, Yih, He, Gao, and Deng]{DistMult}
Bishan Yang, Wen{-}tau Yih, Xiaodong He, Jianfeng Gao, and Li~Deng.
\newblock Embedding entities and relations for learning and inference in
  knowledge bases.
\newblock In \emph{Proceedings of the 3rd International Conference on Learning
  Representations, {ICLR} 2015, San Diego, CA, USA, May 7-9, 2015}, 2015.

\bibitem[Yang et~al.(2017)Yang, Yang, and Cohen]{NeuralLP-NIPS17}
Fan Yang, Zhilin Yang, and William~W. Cohen.
\newblock Differentiable learning of logical rules for knowledge base
  reasoning.
\newblock In \emph{Advances in Neural Information Processing Systems 30: Annual
  Conference on Neural Information Processing Systems 2017, December 4-9, 2017,
  Long Beach, CA, {USA}}, pages 2319--2328, 2017.

\bibitem[Yang et~al.(2022)Yang, Lin, and Zhang]{NIPS22-KG1}
Haotong Yang, Zhouchen Lin, and Muhan Zhang.
\newblock Rethinking knowledge graph evaluation under the open-world
  assumption.
\newblock In \emph{Advances in Neural Information Processing Systems}, 2022.

\bibitem[Zhang et~al.(2019)Zhang, Tay, Yao, and Liu]{QuatE}
Shuai Zhang, Yi~Tay, Lina Yao, and Qi~Liu.
\newblock Quaternion knowledge graph embeddings.
\newblock In \emph{Proceedings of the Annual Conference on Neural Information
  Processing Systems 2019, NeurIPS 2019, December 8-14, 2019, Vancouver, BC,
  Canada}, pages 2731--2741, 2019.

\bibitem[Zhang et~al.(2022{\natexlab{a}})Zhang, Sheng, Yin, Jiang, Xia, Gao,
  Yang, and Cui]{GNNDegrad-KDD22}
Wentao Zhang, Zeang Sheng, Ziqi Yin, Yuezihan Jiang, Yikuan Xia, Jun Gao, Zhi
  Yang, and Bin Cui.
\newblock Model degradation hinders deep graph neural networks.
\newblock In \emph{{KDD} '22: The 28th {ACM} {SIGKDD} Conference on Knowledge
  Discovery and Data Mining, Washington, DC, USA, August 14 - 18, 2022}, pages
  2493--2503. {ACM}, 2022{\natexlab{a}}.

\bibitem[Zhang and Yao(2022)]{REDGNN-WWW22}
Yongqi Zhang and Quanming Yao.
\newblock Knowledge graph reasoning with relational digraph.
\newblock In \emph{{WWW} '22: The {ACM} Web Conference 2022, Virtual Event,
  Lyon, France, April 25 - 29, 2022}, pages 912--924. {ACM}, 2022.

\bibitem[Zhang et~al.(2022{\natexlab{b}})Zhang, Zhou, Yao, Chu, and
  Han]{Adaprop}
Yongqi Zhang, Zhanke Zhou, Quanming Yao, Xiaowen Chu, and Bo~Han.
\newblock Learning adaptive propagation for knowledge graph reasoning.
\newblock \emph{ArXiv}, abs/2205.15319, 2022{\natexlab{b}}.

\bibitem[Zhang et~al.(2021)Zhang, Wang, Chen, Ji, and Wu]{NIPS21-KG2}
Zhanqiu Zhang, Jie Wang, Jiajun Chen, Shuiwang Ji, and Feng Wu.
\newblock Cone: Cone embeddings for multi-hop reasoning over knowledge graphs.
\newblock In \emph{Advances in Neural Information Processing Systems 34: Annual
  Conference on Neural Information Processing Systems 2021, NeurIPS 2021,
  December 6-14, 2021, virtual}, pages 19172--19183, 2021.

\bibitem[Zhu et~al.(2021)Zhu, Zhang, Xhonneux, and Tang]{NBF-NIPS21}
Zhaocheng Zhu, Zuobai Zhang, Louis{-}Pascal A.~C. Xhonneux, and Jian Tang.
\newblock Neural bellman-ford networks: {A} general graph neural network
  framework for link prediction.
\newblock In \emph{Advances in Neural Information Processing Systems 34: Annual
  Conference on Neural Information Processing Systems 2021, NeurIPS 2021,
  December 6-14, 2021, virtual}, pages 29476--29490, 2021.

\bibitem[Zhu et~al.(2022)Zhu, Yuan, Galkin, Xhonneux, Zhang, Gazeau, and
  Tang]{ANet}
Zhaocheng Zhu, Xinyu Yuan, Mikhail Galkin, Sophie Xhonneux, Ming Zhang, Maxime
  Gazeau, and Jian Tang.
\newblock A*net: A scalable path-based reasoning approach for knowledge graphs.
\newblock \emph{ArXiv}, abs/2206.04798, 2022.

\end{thebibliography}
